\documentclass[10pt,twocolumn,letterpaper]{article}

\usepackage[pagenumbers]{cvpr} 

\definecolor{cvprblue}{rgb}{0.21,0.49,0.74}
\usepackage[pagebackref,breaklinks,colorlinks,allcolors=cvprblue]{hyperref}

\usepackage{algorithm}
\usepackage{algorithmic}
\usepackage{amsfonts}
\usepackage{booktabs}
\usepackage{multirow}
\usepackage{amsmath}
\usepackage{textcomp, gensymb}
\usepackage{enumitem}
\usepackage{tcolorbox}
\usepackage{caption}
\DeclareCaptionType{prompt}[Prompt][List of Prompts]

\def\paperID{40492} 
\def\confName{CVPR}
\def\confYear{2026}

\title{SciPostLayoutTree: A Dataset for Structural Analysis of Scientific Posters}

\author{Shohei Tanaka \quad Atsushi Hashimoto \quad Yoshitaka Ushiku \\
OMRON SINIC X Corporation \\
{\tt\small \{shohei.tanaka, atsushi.hashimoto, yoshitaka.ushiku\}@sinicx.com}
}

\begin{document}
\maketitle

\begin{abstract}
Scientific posters play a vital role in academic communication by presenting ideas through visual summaries. Analyzing reading order and parent-child relations of posters is essential for building structure-aware interfaces that facilitate clear and accurate understanding of research content. Despite their prevalence in academic communication, posters remain underexplored in structural analysis research, which has primarily focused on papers. To address this gap, we constructed SciPostLayoutTree, a dataset of approximately 8,000 posters annotated with reading order and parent-child relations. Compared to an existing structural analysis dataset, SciPostLayoutTree contains more instances of spatially challenging relations, including upward, horizontal, and long-distance relations. As a solution to these challenges, we develop Layout Tree Decoder, which incorporates visual features as well as bounding box features including position and category information. The model also uses beam search to predict relations while capturing sequence-level plausibility. Experimental results demonstrate that our model improves the prediction accuracy for spatially challenging relations and establishes a solid baseline for poster structure analysis. The dataset is publicly available at \url{https://huggingface.co/datasets/omron-sinicx/scipostlayouttree}. The code is also publicly available at \url{https://github.com/omron-sinicx/scipostlayouttree}.
\end{abstract}

\section{Introduction}

While the structure of documents and posters is designed to facilitate human understanding, it remains unclear whether ML models can analyze such structure.
Reliable structural analysis is crucial for information access technologies, including structure-aware text-to-speech synthesis~\cite{Xiao2023,hyeon2025mathreadertexttospeechmathematical}, question answering~\cite{tito2023hierarchicalmultimodaltransformersmultipage,vanlandeghem2023documentunderstandingdatasetevaluation}, and retrieval using tables of contents~\cite{li2023structureawarelanguagemodelpretraining}.
It would further aid layout assessment; discrepancies between model predictions and designer intentions could indicate a potentially confusing layout.

Existing work on structural analysis has primarily focused on textual documents, in what is known as document structure analysis (DSA)~\cite{chen2025graphbaseddocumentstructureanalysis,wang2024detectorderconstructtreeconstructionbased}.
However, information is also conveyed through visual media such as scientific posters, a primary medium in academic communication.
Misinterpretation of posters may hinder access to the research content they convey.
This study aims to establish a foundational framework that enables models to interpret and reason over the structural organization of posters.

\begin{figure}[t!]
\centering
\includegraphics[width=\columnwidth]{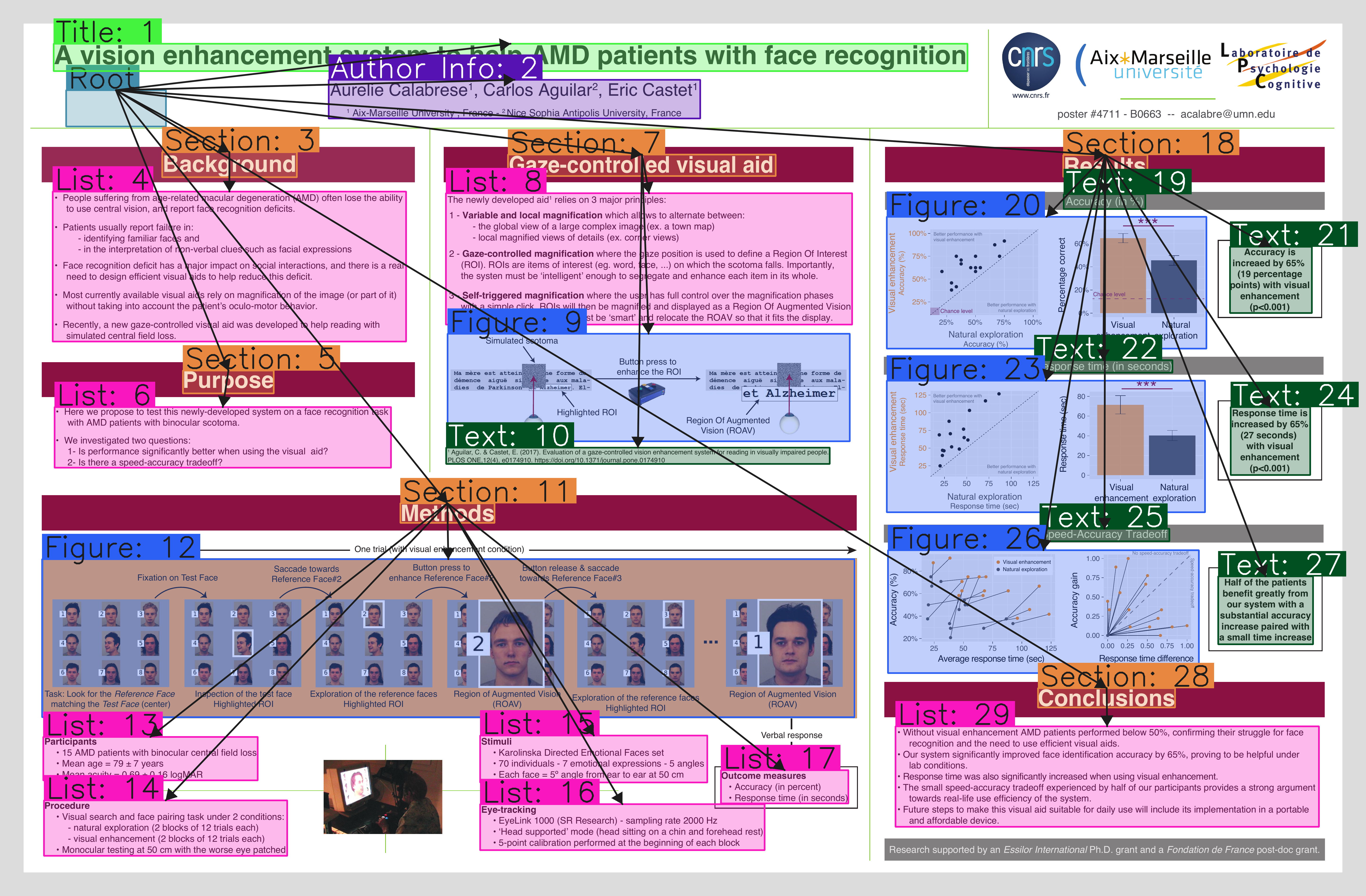}
\caption{
Example from SciPostLayoutTree.
Each arrow denotes a parent-child relation, with the tail indicating the parent and the head indicating the child.
The node labeled ``Root'' denotes the root of the DFS-ordered tree, corresponding to the poster.
The number shown next to each BBox category indicates its reading order priority.
}
\label{fig:dataset_example}
\end{figure}

As a step toward this goal, we construct SciPostLayoutTree, a dataset containing structural annotations for approximately 8,000 posters.
Figure~\ref{fig:dataset_example} shows an example poster with our structure annotation: the reading order and parent-child relations among bounding boxes (BBoxes) are represented as a depth-first-search (DFS)-ordered tree.
A comparative analysis with a DSA dataset reveals that SciPostLayoutTree contains more instances of spatially challenging relations, including upward, horizontal, and long-distance relations.
The result indicates that poster structures present novel technical challenges beyond those found in document structures.

To address these technical challenges, we develop Layout Tree Decoder as an extension of an existing structural analysis model.
The model incorporates not only visual features but also BBox features including position and category information, which enhances its ability to capture spatially challenging relations.
In addition, the model applies beam search during tree decoding to capture sequence-level plausibility.
Experimental results show that our model achieves higher accuracy on spatially challenging relations compared to an existing model using only visual features and greedy decoding.
This observation suggests that our model serves as a solid baseline for poster structure analysis.

Our contributions are summarized as follows:
\begin{itemize}
\item We present SciPostLayoutTree, a novel dataset of approximately 8,000 scientific posters annotated with reading order and parent-child relations, represented as DFS-ordered trees.
\item We show that the posters pose novel spatially challenging relations, including upward, horizontal, and long-distance relations, which are infrequent in documents.
\item We develop Layout Tree Decoder, which provides a solid baseline for poster structure analysis by extending an existing model with BBox features and beam search to capture spatially challenging relations.
\end{itemize}

\section{Related Work}

This section presents existing DSA and scientific poster datasets, contrasting them with our dataset.

\subsection{Document Structure Analysis Dataset}

Existing DSA datasets can be categorized into three types based on their objectives: predicting (i) only reading order~\cite{wang2021layoutreaderpretrainingtextlayout,zhang-etal-2024-modeling,feng2025dolphindocumentimageparsing}, (ii) only hierarchical structure~\cite{ma2023hrdocdatasetbaselinemethod,xing-etal-2024-dochienet}, or (iii) both~\cite{rausch2021docparserhierarchicalstructureparsing,wang2024detectorderconstructtreeconstructionbased,chen2025graphbaseddocumentstructureanalysis}.
These datasets target textual documents and focus on structural analysis challenges such as domain diversity and the prediction of structures spanning multiple pages.
In contrast, our dataset targets scientific posters and is characterized by frequent spatially challenging relations, including upward, horizontal, and long-distance relations.

Among the listed datasets, Comp-HRDoc
~\cite{wang2024detectorderconstructtreeconstructionbased}, GraphDoc~\cite{chen2025graphbaseddocumentstructureanalysis}, and DocHieNet~\cite{xing-etal-2024-dochienet} share similar tasks with our work.
Comp-HRDoc provides annotations for predicting the reading order and parent-child relations of BBoxes as a tree structure.
However, Comp-HRDoc uses text lines as BBox units and focuses on paragraph structuring and section heading detection, whereas our work targets region-level BBoxes such as paragraphs and figures, with different prediction objectives.

GraphDoc provides structural annotations for region-level BBoxes on document pages from DocLayNet~\cite{rausch2021docparserhierarchicalstructureparsing}.
Since GraphDoc is not publicly available, we do not include it in our comparative analysis.
\citeauthor{chen2025graphbaseddocumentstructureanalysis} also developed a model that predicts reading order and parent-child relations between BBoxes based on visual features, which we adopt as a baseline in our experiments.

DocHieNet includes documents of various domains such as scientific papers, financial reports, and legal documents.
We compare DocHieNet with our dataset, as both contain annotations on region-level BBoxes, including reading order and parent-child relations.
However, the model proposed by \citeauthor{xing-etal-2024-dochienet} predicts a hierarchical structure from a given reading order, spanning multiple pages.
This setting is not comparable to our task of jointly predicting reading order and parent-child relations within a single poster.

\subsection{Scientific Poster Dataset}

Although several datasets of scientific posters have been proposed~\cite{1547260908565-217496438,xu2021neural,Tanaka_2024_BMVC,sun2025p2pautomatedpapertopostergeneration}, they focus on poster generation and do not include annotations for structural analysis.
SciPostLayout~\cite{Tanaka_2024_BMVC} contains BBox annotations for approximately 8,000 scientific posters collected from the web.
We construct our dataset by extending SciPostLayout with annotations of reading order and parent-child relations among the BBoxes.

\section{SciPostLayoutTree Dataset}

This section presents the task, the annotation procedure, and the statistical analysis of the dataset.

\subsection{Task Definition}

The DSA task is first defined in Comp-HRDoc~\cite{wang2024detectorderconstructtreeconstructionbased} as predicting a DFS-ordered tree based on reading order and parent-child relations from a set of BBoxes.
We follow this setting for scientific posters.
The input is a set of BBoxes denoted as $B = \{b_1, b_2, \ldots, b_N\}$, where each BBox $b_i$ has center position and size $(x_i, y_i, w_i, h_i)$ as well as category $c_i$.
Here, $N$ denotes the number of BBoxes in the poster. Each BBox is treated as both a layout element and a node in the tree.
In addition, a virtual node $b_0$ representing the poster is introduced as the Root.
The model takes the BBox set $B$ as input and predicts the reading order sequence $\Phi = (\phi(0), \phi(1), \ldots, \phi(N))$.
Here, $\phi(j)$ is a mapping function that outputs the BBox index $i$ corresponding to the 0-based reading order index $j$.
Since the Root node $b_0$ is always placed at the beginning of the reading order, the predicted sequence $\Phi$ is of length $N+1$, starting with $\phi(0) = 0$.
The model also predicts the parent BBox $\psi(i)$ for each BBox $b_i$.
Here, $\psi(i)$ is a mapping function that outputs the index of the parent BBox of $b_i$.
As a result, a DFS-ordered tree $T$ is constructed on the BBox set $B$, where the reading order and parent-child relations are structurally consistent.

\subsection{Dataset Annotation\label{sec:annotation}}

We construct SciPostLayoutTree by adding structural annotations to SciPostLayout~\cite{Tanaka_2024_BMVC}.
In contrast to business or artistic posters, scientific posters have an explicit and consistent reading order intended by the authors.
This property allows us to define a unique DFS-ordered tree structure for each poster.
Each poster was annotated as a DFS-ordered tree rooted at the Root node\footnote{See the Appendix for the annotation guideline.}.
The annotation was conducted by professional annotators from an external vendor, with one annotator assigned per poster.
All annotations were reviewed by in-house supervisors and the authors with expertise in scientific poster layouts, ensuring rigorous quality control.
To quantify annotation consistency, we evaluated 100 randomly selected posters by comparing the original annotations with independently annotated trees by two additional annotators, and observed a high agreement score (0.91)\footnote{See the Appendix for the evaluation details.}.
Posters containing only a single BBox were excluded as structurally trivial.
As a result, we obtained 7,849 annotated posters, split into 6,853/498/498 for training, validation, and test sets.
We confirmed that 98\% of posters in the validation and test sets have non-overlapping authors.
The test set size of 498 posters is sufficient compared to prior datasets such as DocHieNet.


\subsection{Statistical Analysis}

We highlight the unique statistical property of SciPostLayoutTree in comparison with DocHieNet~\cite{xing-etal-2024-dochienet}, which brings us novel technical challenges.

\subsubsection{Layout Element Statistics}

Table~\ref{tab:layout_category} presents statistics on the layout elements in SciPostLayoutTree.
Compared to DocHieNet documents, which contain approximately 12 BBoxes and fewer than one figure per page on average\footnote{These values are derived from A.1 of the DocHieNet paper.}, our posters contain approximately 25 BBoxes and a greater number of figures.
These observations indicate that structural analysis for posters requires predicting relations among a larger number of BBoxes per page than in document-based tasks, and entails greater reliance on visual features.

\begin{table}[t!]
\centering
\small
\caption{
Statistics of layout elements by category.
Mean (Std.) denotes the mean and standard deviation per poster.
}
\begin{tabular}{lrr}
\toprule
Category & Total Count & Mean (Std.) \\
\midrule
Title        &   7,842 &  1.00\ ($\pm$ 0.04) \\
Author Info  &   7,692 &  0.98\ ($\pm$ 0.15) \\
Section      &  41,321 &  5.26\ ($\pm$ 2.43) \\
Text         &  52,725 &  6.72\ ($\pm$ 4.89) \\
List         &  23,761 &  3.03\ ($\pm$ 3.31) \\
Figure       &  38,625 &  4.92\ ($\pm$ 3.62) \\
Table        &   5,655 &  0.72\ ($\pm$ 1.26) \\
Caption      &  14,966 &  1.91\ ($\pm$ 2.66) \\
\midrule
All          & 192,587 & 24.54\ ($\pm$ 9.58) \\
\bottomrule
\end{tabular}
\label{tab:layout_category}
\end{table}

\subsubsection{Layout Tree Statistics}

\begin{figure*}[t!]
\centering
\includegraphics[width=\linewidth]{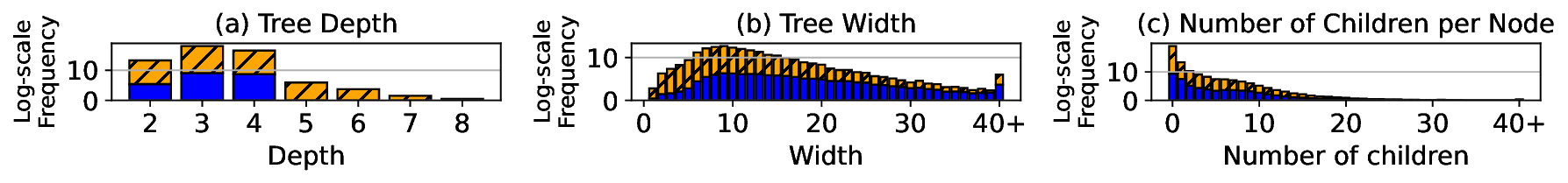}
\caption{
Distributions of tree depth, tree width, and number of children per node.
SciPostLayoutTree (blue) and DocHieNet (orange-hatched) are displayed as stacked bars.
All values are shown on a $\log_2(1 + \text{count})$ scale.
Figures~(a) and (b) show frequencies per 1,000 pages; Figure~(c) shows frequencies per 1,000 nodes.
}
\label{fig:tree_stats_summary}
\end{figure*}

Figure~\ref{fig:tree_stats_summary} summarizes statistics of the overall tree structures in our dataset.
Figures~\ref{fig:tree_stats_summary}~(a)-(b) show that the trees in our dataset are shallower and wider than those in DocHieNet.
Figure~\ref{fig:tree_stats_summary}~(c) shows that the number of children per node follows a skewed distribution, where a small number of nodes have many children.
These characteristics indicate that our dataset contains more cases than DocHieNet in which the model must predict the reading order among many child nodes sharing the same parent.

\subsubsection{Reading Order Statistics}

\begin{figure}[t!]
\centering
\includegraphics[width=\columnwidth]{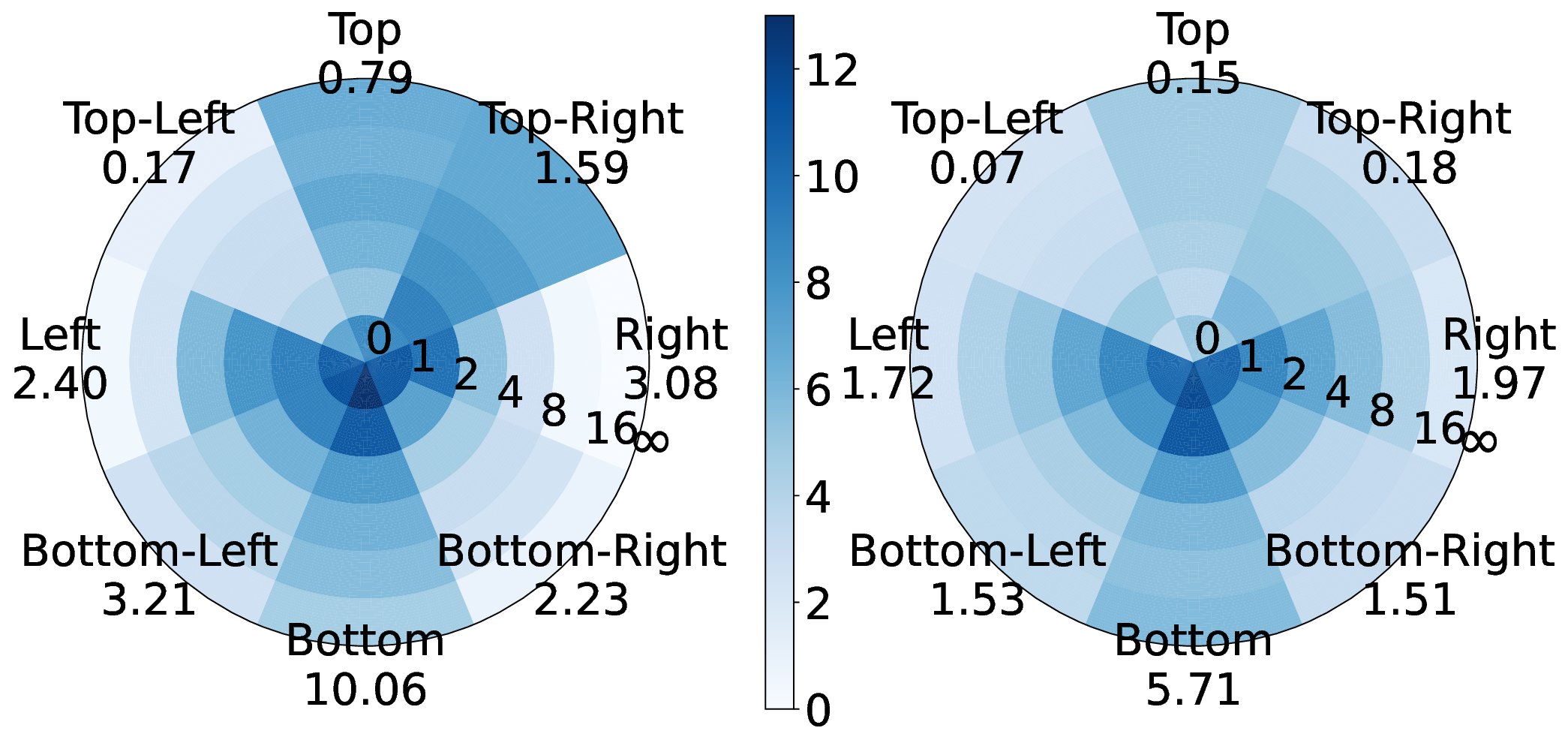}
\caption{
Reading order heatmaps by direction and distance.  
The heatmaps show results from SciPostLayoutTree (left) and DocHieNet (right), with counts normalized per 1,000 pages.  
Distances are binned on a $\log_2$ scale, and heatmap values represent $\log_2(1 + \text{count})$.  
The numbers below each direction indicate the mean count per page.
}
\label{fig:ro_rose_heatmap}
\end{figure}

Figure~\ref{fig:ro_rose_heatmap} uses a polar heatmap to visualize the distribution of directions and distances between consecutive pairs of BBoxes.
Assuming the top-left corner of the image as the origin, we compute the direction class for each consecutive pair of BBoxes with center coordinates $(x_i, y_i)$ and $(x_j, y_j)$ as follows:
\begin{align}
\theta_{i,j} &= \mathrm{arctan2}(y_j - y_i,\, x_j - x_i),\quad \theta_{i,j} \in (-\pi, \pi]\nonumber\\
a_{i,j} &= \left\lfloor \frac{(\theta_{i,j} + 2\pi) \bmod 2\pi + \frac{\pi}{8}}{\frac{\pi}{4}} \right\rfloor \bmod 8 \nonumber
\end{align}
The angle $\theta_{i,j}$ increases in the clockwise direction, with $0$ aligned to the horizontal axis.
The floor function $\lfloor \cdot \rfloor$ returns the greatest integer less than or equal to its argument.
Here, $a_{i,j} \in \{0, \dots, 7\}$ denotes the direction class index in clockwise order, with $0$ corresponding to Right and $7$ to Top-Right.
The distance axis is binned on a $\log_{2}$ scale.
The distance $d$ between BBoxes is defined as follows:
\[
d_{i,j} =
\begin{cases}
\dfrac{|x_j - x_i|}{\max(w_i, w_j)} & \text{if } |x_j - x_i| \geq |y_j - y_i| \\
\dfrac{|y_j - y_i|}{\max(h_i, h_j)} & \text{otherwise}
\end{cases}
\]
\noindent
The distance between BBox centers is normalized by the larger of their widths or heights, depending on the dominant transition axis, to capture scale-invariant patterns.
Normalizing the Euclidean distance by the BBox diagonal may underestimate actual distance, particularly when horizontally elongated BBoxes are vertically arranged.
Color intensity in the heatmap indicates transition frequency in each direction–distance bin, measured on a $\log_{2}$ scale.

The figure shows that, as in DocHieNet, most reading transitions in our dataset are directed downward and span short distances.
However, our dataset includes more transitions toward Top and Top-Right, regardless of distance, and also exhibits frequent horizontal transitions to Right and Left.
These observations indicate that reading order is not constrained to a single direction.

\subsubsection{Parent-Child Relation Statistics}

\begin{figure}[t!]
\centering
\includegraphics[width=\columnwidth]{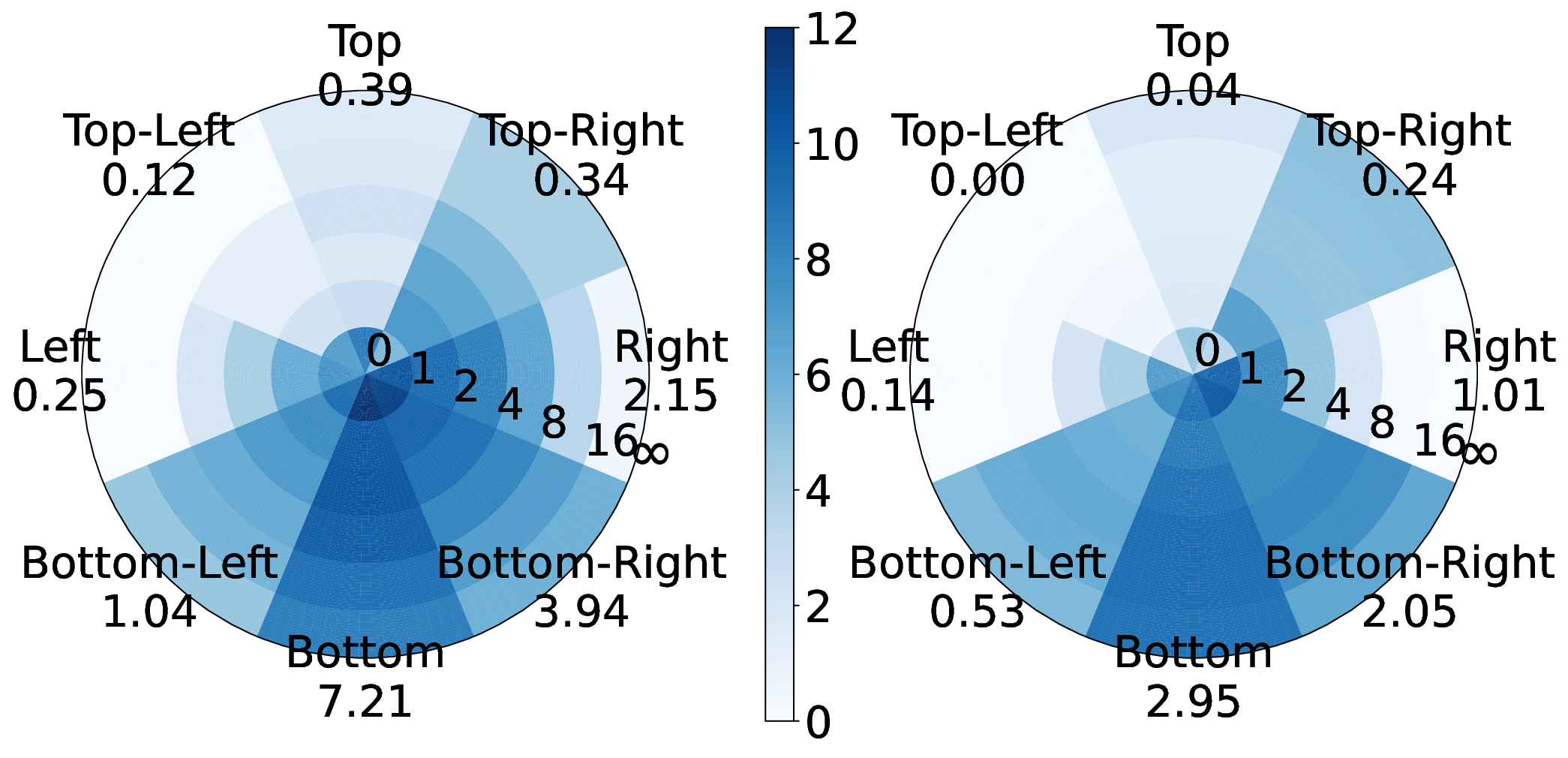}
\caption{
Parent-child relation heatmaps by direction and distance.
The format follows Fig.~\ref{fig:ro_rose_heatmap}.
}
\label{fig:pc_rose_heatmap}
\end{figure}

Figure~\ref{fig:pc_rose_heatmap} presents statistics on parent-child relations.
In our dataset, parent-child relations are primarily directed downward, as in DocHieNet, which suggests that parents are typically positioned above their children.
However, our dataset includes more upward and horizontal relations, indicating increased directional diversity.
Compared to reading order transitions (Fig.~\ref{fig:ro_rose_heatmap}), parent-child relations span longer distances, necessitating the capture of non-local relations for accurate prediction.
Furthermore, as shown in Table~\ref{tab:layout_category}, each poster comprises approximately 25 BBoxes, which necessitates the identification of the correct parent among many candidates.

\section{Layout Tree Decoder}

This section presents the model, the tree decoding algorithm, and the loss function.

\subsection{Model Architecture}

\begin{figure}[t!]
\centering
\includegraphics[width=\columnwidth]{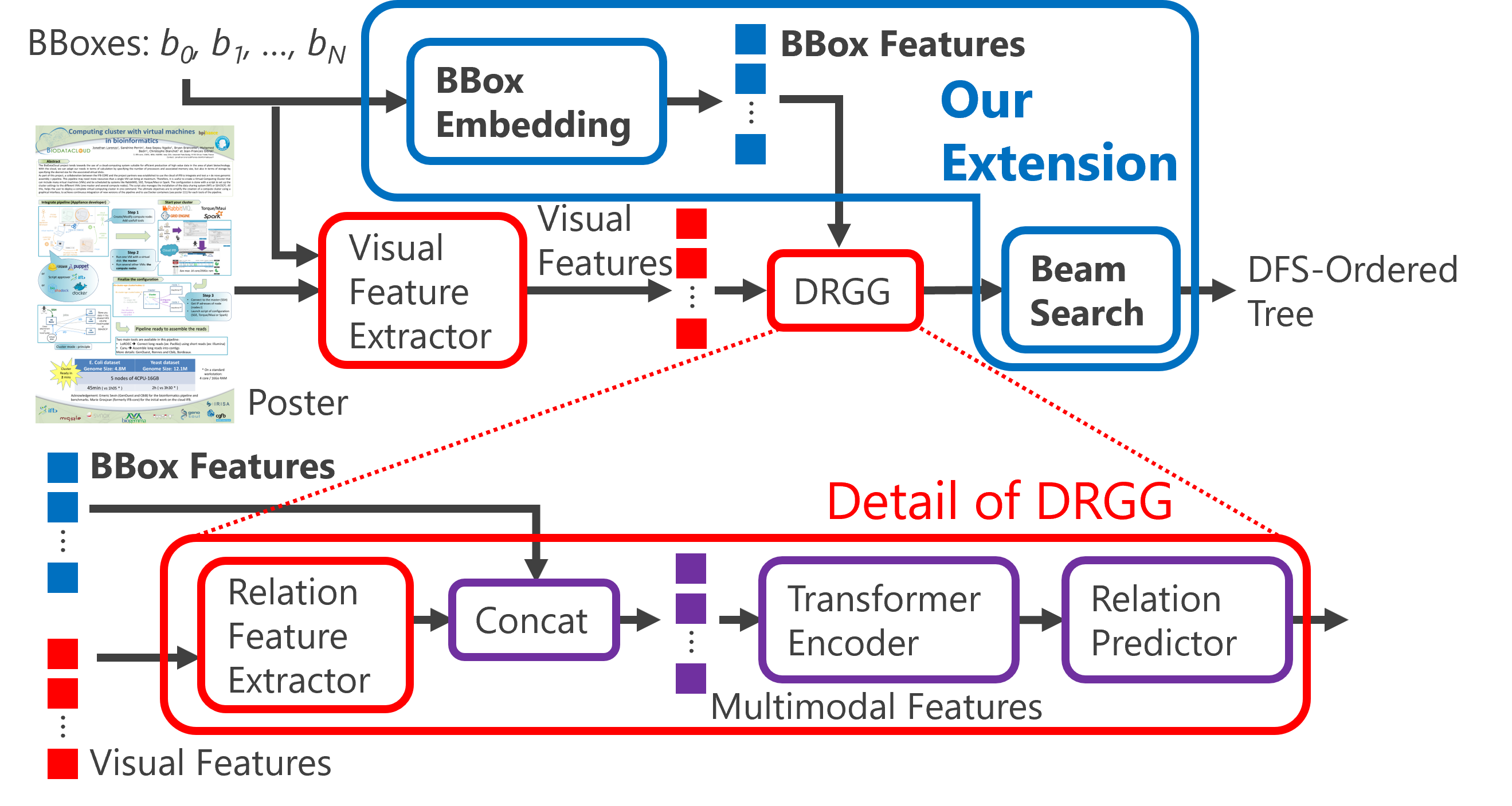}
\caption{
Overview of Layout Tree Decoder.
We extend DRGG by incorporating BBox features and beam search.
}
\label{fig:model}
\end{figure}

As shown in Table~\ref{tab:layout_category}, our dataset contains many figures and thus exhibits strong dependence on visual features.
We therefore build our model on Document Relation Graph Generator (DRGG) proposed by \citeauthor{chen2025graphbaseddocumentstructureanalysis}~(\citeyear{chen2025graphbaseddocumentstructureanalysis}), which predicts relations between BBoxes using visual features.
Figure~\ref{fig:model} presents an overview of our Layout Tree Decoder.
Given an input poster image and BBoxes, the model first extracts a set of visual features $V = \{\mathbf{v_1}, \mathbf{v_2}, \ldots, \mathbf{v_N}\}$ using a Visual Feature Extractor (VFE).
The VFE is constructed by stacking a visual backbone~\cite{he2015deepresiduallearningimage,dosovitskiy2021imageworth16x16words,wang2023internimageexploringlargescalevision}, a Feature Pyramid Network (FPN)~\cite{lin2017featurepyramidnetworksobject}, ROI Align~\cite{he2018maskrcnn}, and a Box Head~\cite{ren2016fasterrcnnrealtimeobject}.
Each vector $\mathbf{v_i}$ represents the visual feature corresponding to the BBox $b_i$.
The feature vector for the Root node $b_0$, denoted as $\mathbf{v_0}$, is obtained by applying a Multilayer Perceptron (MLP) to the FPN output.
We then extend $V$ to include $\mathbf{v_0}$ as $V = \{\mathbf{v_0}, \mathbf{v_1}, \ldots, \mathbf{v_N}\}$.

DRGG applies a Relation Feature Extractor (RFE), which consists of an average pooling layer and MLPs, to the visual features $V$, independently for row and column roles.
The feature sequences used for relation prediction are defined as $V_r = \{\mathbf{v^r_0}, \ldots, \mathbf{v^r_N}\}$ and $V_c = \{\mathbf{v^c_0}, \ldots, \mathbf{v^c_N}\}$, corresponding to the row and column roles, respectively.
$V_r$ represents preceding or child nodes, whereas $V_c$ represents subsequent or parent nodes in a DFS-ordered tree.

Following the DRGG design for visual feature encoding with RFE, we further integrate the spatial and categorical information of each BBox via a BBox Embedding.
For each node $b_i$, the normalized coordinates of the top-left and bottom-right corners are encoded into a position embedding $\mathbf{z_i}$ via an MLP.
The category $c_i$ is embedded as $\mathbf{e_i}$.
For the Root node $b_0$, the coordinate region is fixed to $(0, 0, 1, 1)$, representing the entire poster, and the category is embedded as a special class, ``Root''.
We analyze the effect of combining BBox features with visual features on reading order and parent-child prediction in spatially challenging relations\footnote{Models incorporating OCR text features were also evaluated, but no improvement in prediction accuracy was observed. Details are provided in the Appendix.}.

The BBox features and visual features are combined to obtain the following multimodal features:
\[
\mathbf{m^r_i} = \text{concat}(\mathbf{v^r_i}, \mathbf{z_i}, \mathbf{e_i}),\ 
\mathbf{m^c_i} = \text{concat}(\mathbf{v^c_i}, \mathbf{z_i}, \mathbf{e_i})
\]
\noindent
$\mathbf{m^r_i}$ and $\mathbf{m^c_i}$ are input to a Transformer encoder~\cite{vaswani2023attentionneed} to produce the node representations $F_r$ and $F_c$ for relation prediction.
The relation predictor is an MLP that receives the vector $\text{concat}(F_r[i], F_c[j])$ for each node pair $(b_i, b_j)$ and outputs a score tensor $G \in \mathbb{R}^{2 \times (N+1) \times (N+1)}$.
Here, $G[0]$ and $G[1]$ correspond to the following matrices:
\begin{itemize}
    \item \textbf{Subsequent score matrix} $S \in \mathbb{R}^{(N+1) \times (N+1)}$:
    $S_{i,j}$ indicates the score that $b_j$ follows $b_i$ in the reading order.
    \item \textbf{Parent score matrix} $P \in \mathbb{R}^{(N+1) \times (N+1)}$:  
    $P_{i,j}$ indicates the score that $b_j$ is the parent of $b_i$.
\end{itemize}

\subsection{Tree Decoding Algorithm and Beam Search}

The model decodes a DFS-ordered tree $T$ by applying Algorithm~\ref{alg:tree_decoding} to the score matrices $S$ and $P$.  
This decoding algorithm is derived by reorganizing the reading order and parent-child prediction procedures proposed by \citeauthor{wang2024detectorderconstructtreeconstructionbased}~(\citeyear{wang2024detectorderconstructtreeconstructionbased}).
The decoded tree is consistent with the reading order $\Phi = (\phi(0), \ldots, \phi(N))$ and the parent-child relations.

\begin{algorithm}[t!]
\caption{Tree decoding from score matrices}
\label{alg:tree_decoding}
\begin{algorithmic}[1]
\REQUIRE Subsequent score matrix $S \in \mathbb{R}^{(N+1) \times (N+1)}$, Parent score matrix $P \in \mathbb{R}^{(N+1) \times (N+1)}$
\ENSURE DFS-ordered tree $T$ over $\{0, 1, \ldots, N\}$, where $b_0$ represents the poster and $\{b_1, \ldots, b_N\}$ are its BBoxes
\STATE Initialize $\phi$ as an empty reading order mapping
\STATE Set $\phi(0) \leftarrow 0$
\FOR{$i = 1$ to $N$}
    \STATE Select $\phi(i) \leftarrow \arg\max_{k \notin \{\phi(0), \ldots, \phi(i-1)\}} S_{\phi(i-1),\ k}$
\ENDFOR
\STATE Initialize DFS-ordered tree $T$ with the root node $0$
\STATE Initialize $\psi$ as an empty parent mapping
\STATE Initialize stack $Q \leftarrow [0]$ \COMMENT{The rightmost path from the root node to a leaf node}
\FOR{$j = 1$ to $N$}
    \STATE $i \leftarrow \phi(j)$
    \STATE Select parent $\psi(i) \leftarrow \arg\max_{p \in Q} P_{i, p}$
    \STATE Add node $i$ as the rightmost child of $\psi(i)$ in $T$
    \STATE Truncate $Q$ after $\psi(i)$
    \STATE Push $i$ to $Q$
\ENDFOR
\RETURN $T$
\end{algorithmic}
\end{algorithm}

Previous studies predicted reading order and parent-child relations via greedy decoding~\cite{wang2024detectorderconstructtreeconstructionbased,wang2025unihdsaunifiedrelationprediction}.
However, as shown in Figs.~\ref{fig:ro_rose_heatmap}–\ref{fig:pc_rose_heatmap}, our dataset contains a substantial number of spatially challenging relations, such as upward, horizontal, and long-distance relations.
During greedy decoding, infrequent relations are often not captured due to their low local scores.
To address this issue, we apply beam search to steps 3–5 (reading order prediction) and 9–15 (parent-child prediction) in Algorithm~\ref{alg:tree_decoding} to capture sequential-level plausibility.
Beam search is widely used in natural language processing tasks such as machine translation~\cite{Freitag_2017,vaswani2023attentionneed}, text summarization~\cite{see2017pointsummarizationpointergeneratornetworks}, image captioning~\cite{vinyals2015tellneuralimagecaption}, dialogue generation~\cite{li2016diversitypromotingobjectivefunctionneural}, and named entity recognition~\cite{zhang2023readingordermattersinformation}.
At each decoding step, beam search retains multiple top-ranked candidates to favor sequences with higher total scores despite low local scores.
We evaluate the effect of beam search on reading order and parent-child prediction, particularly for spatially challenging relations.

\subsection{Loss Function}

We define cross-entropy losses based on the subsequent score matrix $S$ and the parent score matrix $P$.
Each BBox $b_i$ has a unique ground-truth (GT) subsequent BBox $b_j^{\text{sub}}$ and a unique parent BBox $b_j^{\text{par}}$.
The losses are defined as follows:
\[
\mathcal{L}_{\text{sub}} = \frac{1}{N} \sum_{i \in \mathcal{I}_{\text{sub}}} \text{CE}(S_i, j^{\text{sub}}_i), \ 
\mathcal{L}_{\text{par}} = \frac{1}{N} \sum_{i \in \mathcal{I}_{\text{par}}} \text{CE}(P_i, j^{\text{par}}_i)
\]
\noindent
Here, $\text{CE}(*)$ denotes the cross-entropy loss for either a score vector $S_i$ or $P_i$, using the GT index $j$ as the label.
The index set $\mathcal{I}_{\text{sub}}$ is $\{0, \phi(1), \ldots, \phi(N-1)\}$, which excludes the terminal node in the reading order sequence.
The index set $\mathcal{I}_{\text{par}}$ is $\{1, 2, \ldots, N\}$, which excludes the Root node.
The total loss is defined as $\mathcal{L} = \mathcal{L}_{\text{sub}} + \mathcal{L}_{\text{par}}$.

\section{Experiments}

This section investigates the effect of incorporating the BBox Embedding and beam search into DRGG.

\subsection{Experimental Settings}

We compared four DRGG variants, each with a different combination of the BBox Embedding and beam search~\footnote{We also evaluated VLMs (GPT-5, Gemini 3 Pro), but they showed lower performance than DRGG-based models. Details are provided in the Appendix.}.

\noindent
\textbf{DRGG}
predicts score matrices between BBoxes based on the visual features, and determines the reading order and parent-child relations through greedy decoding.

\noindent
\textbf{DRGG-BE (BBox Embedding)}
augments the input features with the BBox features.

\noindent
\textbf{DRGG-BS (Beam Search)}
replaces greedy decoding with beam search to predict the relations.

\noindent
\textbf{DRGG-BEBS (BBox Embedding and Beam Search)}
integrates DRGG-BE and DRGG-BS.

To confirm that performance differences among methods are not attributable to backbone choice, we employed five visual backbones: ResNet-50~\cite{he2015deepresiduallearningimage}, ViT~\cite{dosovitskiy2021imageworth16x16words}, Swin~\cite{liu2021swintransformerhierarchicalvision}, DiT~\cite{li2022ditselfsupervisedpretrainingdocument}, and InternImage~\cite{wang2023internimageexploringlargescalevision}.
The beam width was set to 20 for DRGG-BS and DRGG-BEBS\footnote{See the Appendix for experimental details.}.

\subsection{Evaluation Metrics\label{sec:metrics}}

Three metrics were used to evaluate the predicted DFS-ordered trees: Tree Edit Distance (TED)~\cite{PAWLIK2016157,10.1145/2699485}, Semantic Tree Edit Distance Score (STEDS)~\cite{ma2023hrdocdatasetbaselinemethod}, and Reading Edit Distance Score (REDS)~\cite{wang2024detectorderconstructtreeconstructionbased}.
TED is the edit distance between the predicted tree $T_P$ and the GT tree $T_G$, defined as the minimum number of node insertions, deletions, and substitutions.
STEDS is a normalized version of TED with respect to tree size, defined as follows:
\[
\text{STEDS} = 100\left(1 - \frac{\text{TED}(T_G, T_P)}{\max(|T_G|, |T_P|)}\right)
\]
\noindent
Here, $|T_G|$ and $|T_P|$ denote the number of nodes in the GT tree and the predicted tree.
REDS evaluates the similarity between reading orders using the Levenshtein distance (string edit distance) $\text{LD}(\Phi_G, \Phi_P)$ between the GT order $\Phi_G$ and the predicted order $\Phi_P$.
The metric is defined as follows:
\[
\text{REDS} = 100\left(1 - \frac{\text{LD}(\Phi_G, \Phi_P)}{\max(|T_G|, |T_P|)}\right)
\]

The accuracies were computed for each direction and distance.
These accuracies are defined as the ratio of GT relations included in the predicted tree.
Relations involving the Root were excluded from the accuracy computation, as direction and distance are undefined for them.

\begin{table}[t!]
\centering
\small
\caption{
Comparison of DRGG and its extensions across visual backbones.
${\ast}$ indicates significant improvement over DRGG at $p < 0.005$ according to Wilcoxon signed-rank test.
}
\setlength{\tabcolsep}{2pt}
\begin{tabular}{ll|rrr}
\toprule
Backbone & Decoder & STEDS ($\uparrow$) & REDS ($\uparrow$) & TED ($\downarrow$) \\
\midrule
\multirow{4}{*}{ResNet-50} & DRGG & 68.74 & 75.07 & 8.83 \\
    & DRGG-BE & 84.24$^{\ast}$ & 86.44$^{\ast}$ & 4.41$^{\ast}$ \\
    & DRGG-BS & 76.79$^{\ast}$ & 83.14$^{\ast}$ & 6.65$^{\ast}$ \\
    & DRGG-BEBS & 88.45$^{\ast}$ & 90.40$^{\ast}$ & 3.22$^{\ast}$ \\
\midrule
\multirow{4}{*}{ViT} & DRGG & 80.40 & 85.94 & 5.45 \\
    & DRGG-BE & 86.38$^{\ast}$ & 88.24$^{\ast}$ & 3.85$^{\ast}$ \\
    & DRGG-BS & 83.89$^{\ast}$ & 89.46$^{\ast}$ & 4.46$^{\ast}$ \\
    & DRGG-BEBS & 90.04$^{\ast}$ & 91.73$^{\ast}$ & 2.78$^{\ast}$ \\
\midrule
\multirow{4}{*}{Swin} & DRGG & 79.90 & 85.77 & 5.69 \\
    & DRGG-BE & 86.73$^{\ast}$ & 88.42$^{\ast}$ & 3.69$^{\ast}$ \\
    & DRGG-BS & 83.11$^{\ast}$ & 89.02$^{\ast}$ & 4.74$^{\ast}$ \\
    & DRGG-BEBS & 89.26$^{\ast}$ & 90.88$^{\ast}$ & 2.95$^{\ast}$ \\
\midrule
\multirow{4}{*}{DiT} & DRGG & 78.33 & 84.81 & 6.09 \\
    & DRGG-BE & 85.32$^{\ast}$ & 87.36$^{\ast}$ & 4.22$^{\ast}$ \\
    & DRGG-BS & 82.12$^{\ast}$ & 88.72$^{\ast}$ & 5.07$^{\ast}$ \\
    & DRGG-BEBS & 88.69$^{\ast}$ & 90.63$^{\ast}$ & 3.24$^{\ast}$ \\
\midrule
\multirow{4}{*}{InternImage} & DRGG & 80.19 & 85.53 & 5.53 \\
    & DRGG-BE & 86.89$^{\ast}$ & 88.56$^{\ast}$ & 3.70$^{\ast}$ \\
    & DRGG-BS & 83.75$^{\ast}$ & 89.25$^{\ast}$ & 4.55$^{\ast}$ \\
    & DRGG-BEBS & 89.34$^{\ast}$ & 90.97$^{\ast}$ & 2.93$^{\ast}$ \\
\bottomrule
\end{tabular}
\label{tab:main_results}
\end{table}

\begin{table*}[ht!]
\centering
\small
\caption{
Per-direction accuracy of the reading order prediction with DRGG-BEBS.
Values in parentheses indicate the accuracy improvement over DRGG-BE.
${\star}$ and ${\star\star}$ indicate significant improvements over DRGG-BE at $p < 0.005$ and $p < 0.05$, respectively, according to McNemar's test.
}
\setlength{\tabcolsep}{5pt}
\begin{tabular}{l|rrrrrrrr}
\toprule
\multicolumn{1}{l}{Backbone} & \multicolumn{1}{|l}{Right} & \multicolumn{1}{l}{Bottom-Right} & \multicolumn{1}{l}{Bottom} & \multicolumn{1}{l}{Bottom-Left} & \multicolumn{1}{l}{Left} & \multicolumn{1}{l}{Top-Left} & \multicolumn{1}{l}{Top} & \multicolumn{1}{l}{Top-Right} \\
\midrule
ResNet-50       & 86.0\ (+4.0)$^{\star}$ & 94.5\ (+0.6) & 96.7\ (+2.0)$^{\star}$ & 92.7\ (+2.4)$^{\star}$ & 84.2\ (+3.6)$^{\star}$ & 73.3\ (-1.7) & 87.5\ (+2.9) & 79.8\ (+5.5)$^{\star}$ \\
ViT             & 87.7\ (+2.8)$^{\star}$ & 95.2\ (+0.0) & 97.2\ (+1.3)$^{\star}$ & 93.1\ (+2.0)$^{\star}$ & 85.4\ (+2.6)$^{\star}$ & 75.0\ (+5.0) & 89.3\ (+2.3) & 81.4\ (+5.7)$^{\star}$ \\
Swin            & 87.9\ (+2.4)$^{\star}$ & 95.0\ (-0.5) & 97.0\ (+1.1)$^{\star}$ & 92.7\ (+1.0)$^{\star\star}$ & 85.5\ (+2.8)$^{\star}$ & 73.3\ (-5.0) & 89.0\ (+2.0) & 81.1\ (+3.7)$^{\star}$ \\
DiT             & 85.3\ (+2.8)$^{\star}$ & 94.6\ (+0.5) & 96.6\ (+1.6)$^{\star}$ & 92.9\ (+2.7)$^{\star}$ & 83.3\ (+3.3)$^{\star}$ & 71.7\ (-3.3) & 88.1\ (+2.3) & 79.3\ (+5.5)$^{\star}$ \\
InternImage     & 87.5\ (+2.1)$^{\star}$ & 94.8\ (+0.0) & 96.4\ (+0.7)$^{\star}$ & 92.9\ (+1.3)$^{\star}$ & 85.9\ (+2.9)$^{\star}$ & 73.3\ (+3.3) & 88.1\ (+2.3)$^{\star\star}$ & 80.1\ (+3.0)$^{\star}$ \\
\bottomrule
\end{tabular}
\label{tab:ro_direction_accuracy}
\end{table*}

\begin{table*}[ht!]
\centering
\small
\caption{
Per-distance accuracy of the reading order prediction.
The format and experimental settings follow Table~\ref{tab:ro_direction_accuracy}.
}
\begin{tabular}{l|rrrrrr}
\toprule
Backbone & (0, 1] & (1, 2] & (2, 4] & (4, 8] & (8, 16] & (16, $\infty$) \\
\midrule
ResNet-50     & 94.7\ (+1.9)$^{\star}$ & 84.0\ (+4.1)$^{\star}$ & 77.4\ (+6.4)$^{\star}$ & 85.4\ (+5.6)$^{\star\star}$ & 86.3\ (+5.3)$^{\star\star}$ & 89.4\ (+3.3) \\
ViT           & 95.3\ (+1.3)$^{\star}$ & 85.5\ (+3.1)$^{\star}$ & 80.5\ (+4.4)$^{\star}$ & 86.4\ (+8.1)$^{\star}$ & 87.0\ (+2.3) & 89.4\ (+4.9)$^{\star\star}$ \\
Swin          & 95.2\ (+1.0)$^{\star}$ & 85.5\ (+2.6)$^{\star}$ & 78.7\ (+2.2) & 84.3\ (+3.0) & 87.0\ (+1.5) & 90.2\ (+4.9)$^{\star\star}$ \\
DiT           & 94.4\ (+1.6)$^{\star}$ & 84.1\ (+3.8)$^{\star}$ & 77.2\ (+5.1)$^{\star}$ & 83.3\ (+3.5) & 87.8\ (+3.8) & 91.1\ (+0.8) \\
InternImage   & 94.8\ (+0.9)$^{\star}$ & 85.1\ (+2.1)$^{\star}$ & 78.0\ (+4.0)$^{\star}$ & 88.9\ (+2.5) & 86.3\ (+2.3) & 87.8\ (+3.3) \\
\bottomrule
\end{tabular}
\label{tab:ro_distance_accuracy}
\end{table*}

\begin{table*}[ht!]
\centering
\small
\caption{
Per-direction accuracy of the parent-child prediction with DRGG-BEBS.
Values in parentheses indicate the accuracy improvement over DRGG-BS.
${\star}$ and ${\star\star}$ indicate significant improvements over DRGG-BS at $p < 0.005$ and $p < 0.05$, respectively, according to McNemar's test.
}
\setlength{\tabcolsep}{3pt}
\begin{tabular}{l|rrrrrrrr}
\toprule
\multicolumn{1}{l}{Backbone} & \multicolumn{1}{|l}{Right} & \multicolumn{1}{l}{Bottom-Right} & \multicolumn{1}{l}{Bottom} & \multicolumn{1}{l}{Bottom-Left} & \multicolumn{1}{l}{Left} & \multicolumn{1}{l}{Top-Left} & \multicolumn{1}{l}{Top} & \multicolumn{1}{l}{Top-Right} \\
\midrule
ResNet-50       & 93.3\ (+13.3)$^{\star}$ & 95.3\ (+8.4)$^{\star}$ & 98.0\ (+9.2)$^{\star}$ & 92.6\ (+13.6)$^{\star}$ & 74.5\ (+19.1)$^{\star}$ & 90.7 (+18.6)$^{\star\star}$ & 88.8\ (+14.0)$^{\star}$ & 83.4 (+10.4)$^{\star\star}$ \\
ViT             & 94.3\ (+5.3)$^{\star}$ & 97.1\ (+3.7)$^{\star}$ & 98.4\ (+5.5)$^{\star}$ & 94.7\ (+9.3)$^{\star}$ & 77.3 (+7.1) & 86.0 (+7.0) & 89.9\ (+20.8)$^{\star}$ & 87.6\ (+12.4)$^{\star}$ \\
Swin            & 94.8\ (+7.9)$^{\star}$ & 97.1\ (+6.1)$^{\star}$ & 98.5\ (+6.6)$^{\star}$ & 92.1\ (+8.5)$^{\star}$ & 73.0\ (+9.9)$^{\star}$ & 88.4 (+14.0)$^{\star\star}$ & 91.0\ (+21.3)$^{\star}$ & 90.2\ (+14.0)$^{\star}$ \\
DiT             & 92.8\ (+7.7)$^{\star}$ & 96.3\ (+6.5)$^{\star}$ & 97.8\ (+7.2)$^{\star}$ & 94.7\ (+12.1)$^{\star}$ & 80.1\ (+14.9)$^{\star}$ & 83.7 (+23.3)$^{\star\star}$ & 90.4\ (+32.6)$^{\star}$ & 84.5\ (+9.3)$^{\star}$ \\
InternImage     & 93.8\ (+6.1)$^{\star}$ & 96.9\ (+5.5)$^{\star}$ & 98.5\ (+6.1)$^{\star}$ & 94.3\ (+10.8)$^{\star}$ & 75.9 (+7.1)$^{\star\star}$ & 88.4 (+11.6) & 90.4\ (+13.5)$^{\star}$ & 92.2\ (+16.6)$^{\star}$ \\
\bottomrule
\end{tabular}
\label{tab:pc_direction_accuracy}
\end{table*}

\begin{table*}[ht!]
\centering
\small
\caption{
Per-distance accuracy of the parent-child prediction.
The format and experimental settings follow Table~\ref{tab:pc_direction_accuracy}.
}
\begin{tabular}{l|rrrrrr}
\toprule
Backbone & (0, 1] & (1, 2] & (2, 4] & (4, 8] & (8, 16] & (16, $\infty$) \\
\midrule
ResNet-50       & 97.8\ (+6.4)$^{\star}$ & 92.2\ (+13.1)$^{\star}$ & 92.0\ (+14.7)$^{\star}$ & 92.3\ (+13.3)$^{\star}$ & 91.1\ (+16.5)$^{\star}$ & 93.7\ (+28.3)$^{\star}$ \\
ViT             & 98.3\ (+5.1)$^{\star}$ & 92.8\ (+4.2)$^{\star}$ & 94.8\ (+6.6)$^{\star}$ & 94.9\ (+5.7)$^{\star}$ & 94.0\ (+9.5)$^{\star}$ & 95.6\ (+23.4)$^{\star}$ \\
Swin            & 98.5\ (+5.8)$^{\star}$ & 93.4\ (+7.6)$^{\star}$ & 94.2\ (+9.0)$^{\star}$ & 93.5\ (+7.9)$^{\star}$ & 93.4\ (+12.3)$^{\star}$ & 94.6\ (+20.5)$^{\star}$ \\
DiT             & 97.9\ (+7.2)$^{\star}$ & 92.9\ (+7.6)$^{\star}$ & 92.4\ (+9.5)$^{\star}$ & 93.1\ (+7.7)$^{\star}$ & 93.0\ (+15.2)$^{\star}$ & 94.6\ (+20.0)$^{\star}$ \\
InternImage     & 98.4\ (+5.2)$^{\star}$ & 93.1\ (+7.1)$^{\star}$ & 93.5\ (+8.0)$^{\star}$ & 94.6\ (+7.5)$^{\star}$ & 95.6\ (+13.3)$^{\star}$ & 95.6\ (+17.6)$^{\star}$ \\
\bottomrule
\end{tabular}
\label{tab:pc_distance_accuracy}
\end{table*}

\subsection{Evaluation Results}

Table~\ref{tab:main_results} shows that both the BBox Embedding and beam search consistently improve STEDS, REDS, and TED.
The BBox Embedding improves both STEDS and REDS, with a more substantial effect on STEDS.
The results suggest that incorporating the BBox Embedding is effective for improving the accuracy of the parent-child prediction.

Beam search contributes to improvements in REDS, while its effect on STEDS is less substantial than that of the BBox Embedding.
This difference reflects the structural properties of the two prediction tasks.
Reading order prediction is sequential, with each decision depending on the previous decisions.  
In contrast, parent-child prediction does not depend on previous decisions, except when constructing the rightmost path.
Beam search biases the decoding process toward reading orders with higher sequence-level plausibility.

DRGG-BEBS, which integrates the BBox Embedding and beam search, achieves the highest scores across all metrics.
This result suggests that the two components are complementary.
Furthermore, these improvements are consistent across visual backbones, demonstrating that the effectiveness of our extension is not dependent on the backbone.

Despite these improvements, some posters remain challenging, as illustrated in Figure~\ref{fig:error_example}\footnote{See the Appendix for additional challenging examples.}.
In this poster, all Figures and Tables are attached as children to the bottom-left Section, following a Z-shaped reading order.
Accurate prediction of such relations requires capturing semantic grouping and structural plausibility over longer sequences.

\begin{figure}[t!]
\centering
\includegraphics[width=0.9\columnwidth]{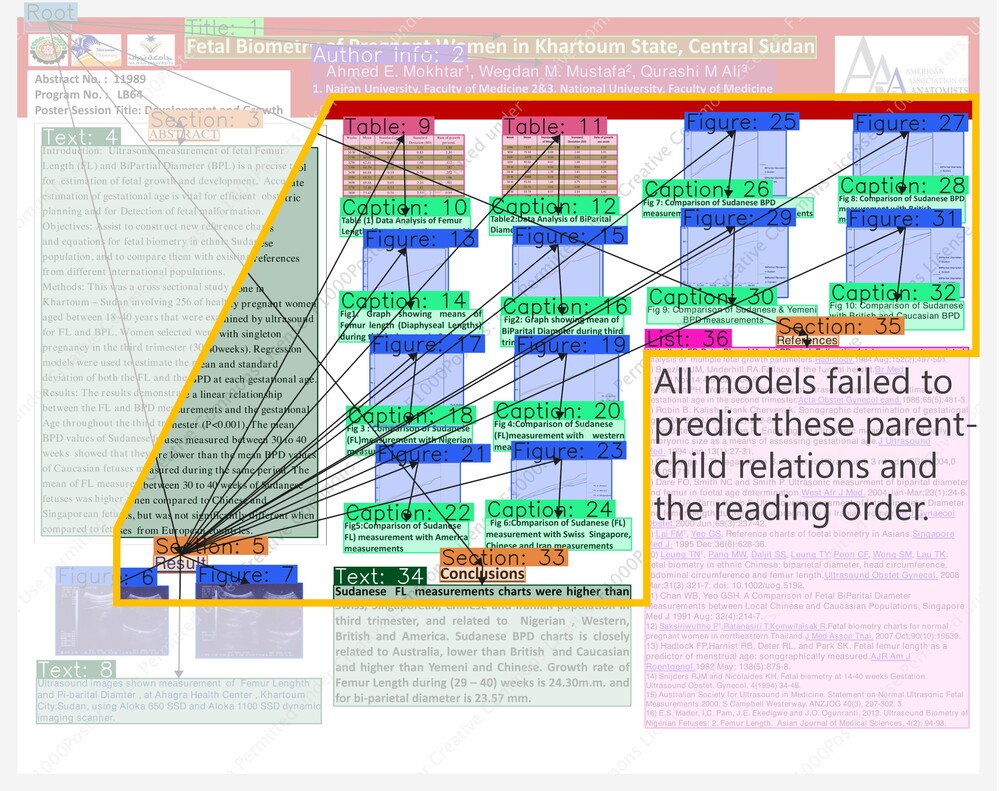}
\caption{
Example of a poster with the GT annotation, where the predicted tree receives a low STEDS (42.70).
}
\label{fig:error_example}
\end{figure}

\subsubsection{Analysis of Reading Order Prediction Improvements by Beam Search}

The directional analysis in Table~\ref{tab:ro_direction_accuracy} shows that downward relations have high baseline accuracy, which limits the improvement obtained through beam search\footnote{See the Appendix for results on all backbones and decoders.\label{ft:danalysis}}.
In contrast, upward and horizontal relations have lower baseline accuracy, and beam search leads to substantial improvements in these cases.
The distance-based analysis in Table~\ref{tab:ro_distance_accuracy} reveals a similar pattern.
Short-distance relations show high baseline accuracy, resulting in limited improvement, whereas long-distance relations benefit significantly from beam search.
These results suggest that beam search mitigates the bias toward downward and short-distance relations, allowing the model to predict correct reading orders even when relations are spatially challenging.

\begin{figure}[t!]
\centering
\includegraphics[width=0.9\columnwidth]{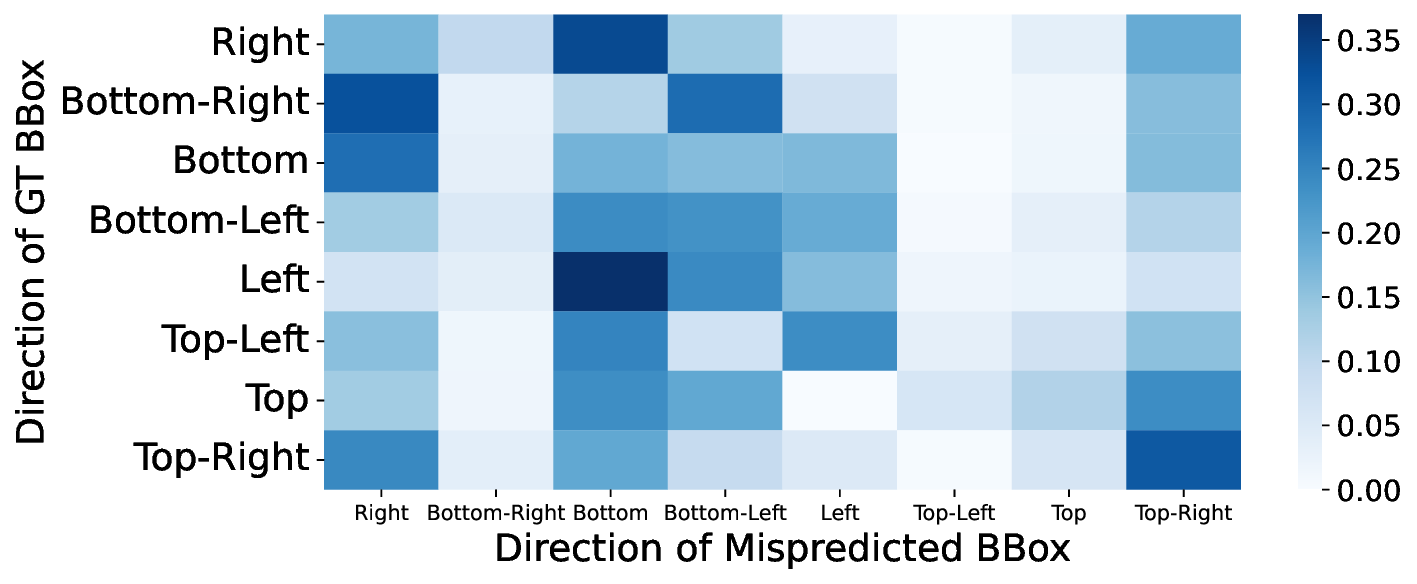}
\caption{
Distribution of reading order prediction failures for each GT BBox direction.
Failures from all backbones using DRGG-BEBS are accumulated in this visualization.
These failures account for 7.00\% of all predictions.
}
\label{fig:ro_error_direction}
\end{figure}

\begin{figure}[t!]
\centering
\includegraphics[width=0.9\columnwidth]{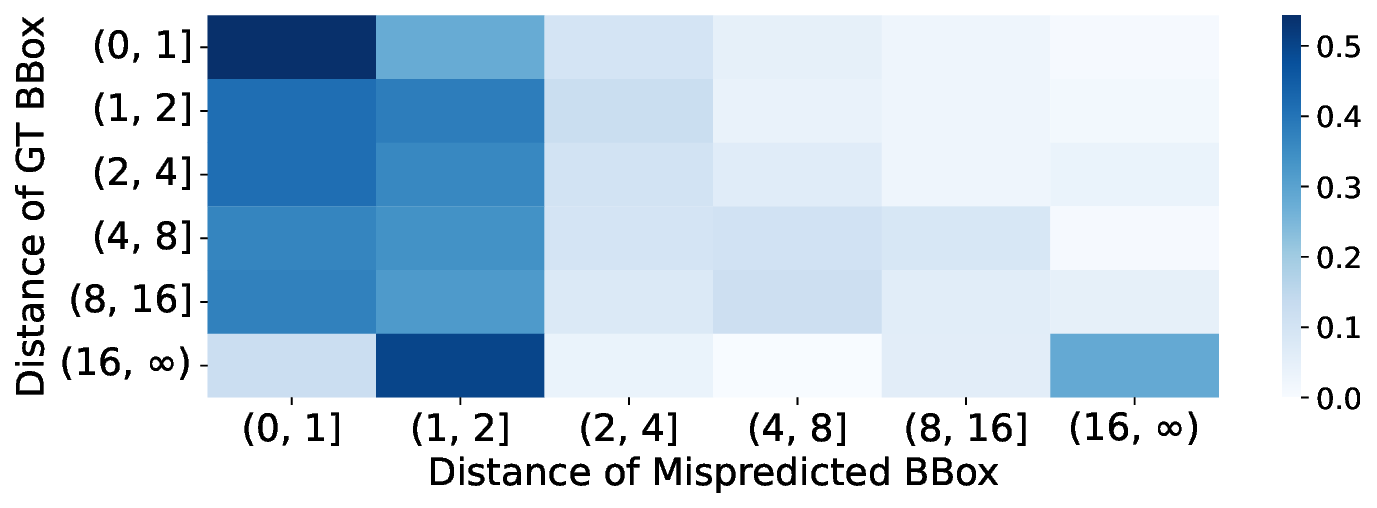}
\caption{
Distribution of reading order prediction failures for each GT BBox distance.
These failures account for 7.00\% of all predictions.
The format follows Fig.~\ref{fig:ro_error_direction}.
}
\label{fig:ro_error_distance}
\end{figure}

However, the error analyses in Figures~\ref{fig:ro_error_direction}--\ref{fig:ro_error_distance} revealed that even the models with beam search frequently mispredicted reading orders biased toward downward and short-distance relations.
These limitations indicate that further progress in reading order prediction requires decoding strategies that reduce systematic directional and distance biases.

\subsubsection{Analysis of Parent-Child Prediction Improvements by BBox Embedding}

Tables~\ref{tab:pc_direction_accuracy}–\ref{tab:pc_distance_accuracy} show that accuracy for spatially challenging relations improved more than that for downward or short-distance relations\footref{ft:danalysis}.
These results indicate the effectiveness of BBox information in improving performance on such challenging cases.

\begin{figure}[t!]
\centering
\includegraphics[width=0.9\columnwidth]{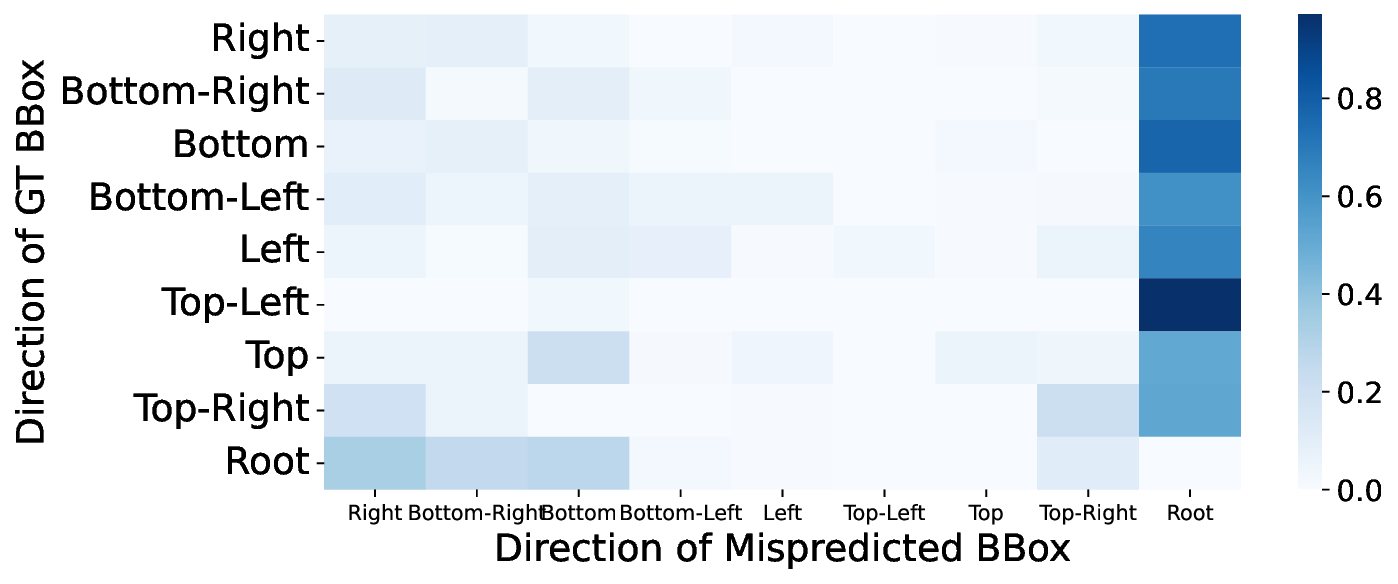}
\caption{
Distribution of parent-child prediction failures for each GT BBox direction.
These failures account for 3.59\% of all predictions.
The format follows Fig.~\ref{fig:ro_error_direction}.
}
\label{fig:pc_error_direction}
\end{figure}

\begin{figure}[t!]
\centering
\includegraphics[width=0.9\columnwidth]{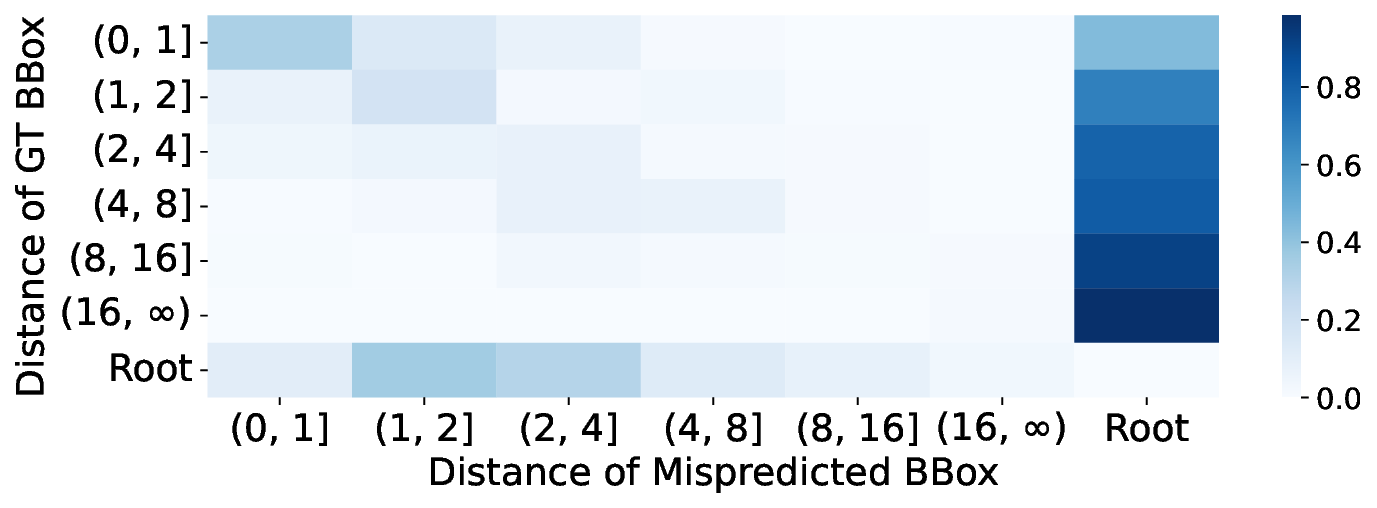}
\caption{
Distribution of parent-child prediction failures for each GT BBox distance.
These failures account for 3.59\% of all predictions.
The format follows Fig.~\ref{fig:ro_error_direction}.
}
\label{fig:pc_error_distance}
\end{figure}

\begin{figure}[t!]
\centering
\includegraphics[width=0.9\columnwidth]{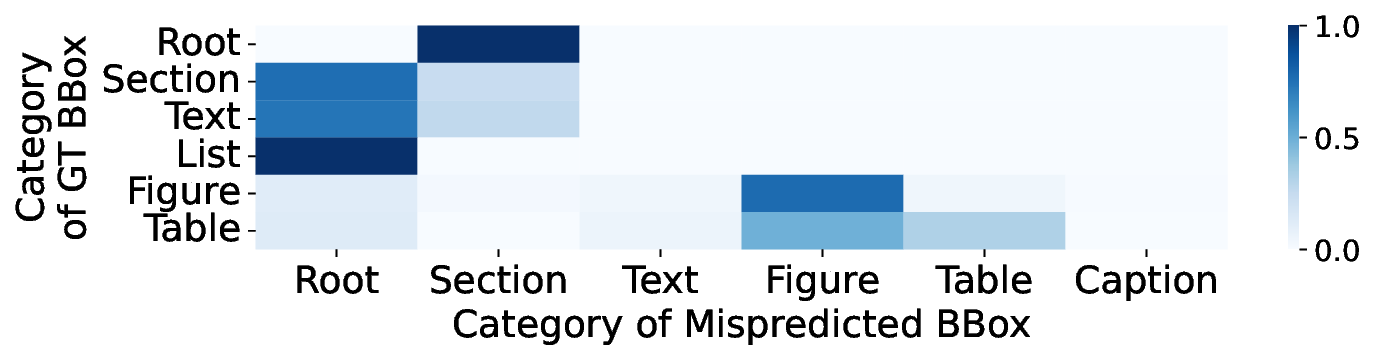}
\caption{
Distribution of parent-child prediction failures for each GT BBox category.
These failures account for 3.59\% of all predictions.
The format follows Fig.~\ref{fig:ro_error_direction}.
}
\label{fig:pc_error_category}
\end{figure}

In contrast, the error analysis in Figures~\ref{fig:pc_error_direction}--\ref{fig:pc_error_distance} revealed that the Root node was frequently mispredicted as the parent, regardless of the correct parent’s direction or distance.
Figure~\ref{fig:pc_error_category} shows that most of these errors involved nodes such as Text, List, Figure, or Table, which should have had Section as their parent.
Our analysis revealed that in 60\% of the cases, the correct Section node was present in the constructed tree but not on the rightmost path, and was therefore excluded from the selectable parent candidates.
These errors arise from incorrect predictions in earlier steps and are difficult to resolve in subsequent decoding.
The observed limitations indicate that progress in parent-child prediction depends on a model architecture capable of capturing not only pairwise relations but also the poster’s structural organization.

\section{Conclusion}

We constructed SciPostLayoutTree, a dataset of approximately 8,000 scientific posters annotated with reading order and parent-child relations in the form of DFS-ordered trees.
This dataset contains more instances of spatially challenging relations, such as upward, horizontal, and long-distance relations, compared to documents.
We extended a relation prediction model that uses visual features and greedy decoding by incorporating the BBox Embedding and beam search.
Experimental results show that the BBox Embedding improves the accuracy of spatially challenging parent-child prediction, while beam search improves the accuracy of spatially challenging reading order prediction.
Furthermore, error analysis revealed the importance of developing models that better capture the structural organization of posters while mitigating bias toward downward and short-distance relations.

\paragraph{Acknowledgments.}
This work was supported by the JST Moonshot R\&D Program (Grant Number JPMJMS2236) and by JSPS KAKENHI (Grant Number 25H01149).


{
    \small
    \bibliographystyle{ieeenat_fullname}
    \bibliography{main}

\begin{thebibliography}{38}
\providecommand{\natexlab}[1]{#1}
\providecommand{\url}[1]{\texttt{#1}}
\expandafter\ifx\csname urlstyle\endcsname\relax
  \providecommand{\doi}[1]{doi: #1}\else
  \providecommand{\doi}{doi: \begingroup \urlstyle{rm}\Url}\fi

\bibitem[Beltagy et~al.(2019)Beltagy, Lo, and Cohan]{beltagy2019scibertpretrainedlanguagemodel}
Iz Beltagy, Kyle Lo, and Arman Cohan.
\newblock Scibert: A pretrained language model for scientific text, 2019.

\bibitem[Chen et~al.(2025)Chen, Liu, Zheng, Wen, Peng, Zhang, and Stiefelhagen]{chen2025graphbaseddocumentstructureanalysis}
Yufan Chen, Ruiping Liu, Junwei Zheng, Di Wen, Kunyu Peng, Jiaming Zhang, and Rainer Stiefelhagen.
\newblock Graph-based document structure analysis, 2025.

\bibitem[Dosovitskiy et~al.(2021)Dosovitskiy, Beyer, Kolesnikov, Weissenborn, Zhai, Unterthiner, Dehghani, Minderer, Heigold, Gelly, Uszkoreit, and Houlsby]{dosovitskiy2021imageworth16x16words}
Alexey Dosovitskiy, Lucas Beyer, Alexander Kolesnikov, Dirk Weissenborn, Xiaohua Zhai, Thomas Unterthiner, Mostafa Dehghani, Matthias Minderer, Georg Heigold, Sylvain Gelly, Jakob Uszkoreit, and Neil Houlsby.
\newblock An image is worth 16x16 words: Transformers for image recognition at scale, 2021.

\bibitem[Feng et~al.(2025)Feng, Wei, Fei, Shi, Han, Liao, Lu, Wu, Liu, Lin, Tang, Liu, and Huang]{feng2025dolphindocumentimageparsing}
Hao Feng, Shu Wei, Xiang Fei, Wei Shi, Yingdong Han, Lei Liao, Jinghui Lu, Binghong Wu, Qi Liu, Chunhui Lin, Jingqun Tang, Hao Liu, and Can Huang.
\newblock Dolphin: Document image parsing via heterogeneous anchor prompting, 2025.

\bibitem[Freitag and Al-Onaizan(2017)]{Freitag_2017}
Markus Freitag and Yaser Al-Onaizan.
\newblock Beam search strategies for neural machine translation.
\newblock In \emph{Proceedings of the First Workshop on Neural Machine Translation}. Association for Computational Linguistics, 2017.

\bibitem[He et~al.(2015)He, Zhang, Ren, and Sun]{he2015deepresiduallearningimage}
Kaiming He, Xiangyu Zhang, Shaoqing Ren, and Jian Sun.
\newblock Deep residual learning for image recognition, 2015.

\bibitem[He et~al.(2018)He, Gkioxari, Dollár, and Girshick]{he2018maskrcnn}
Kaiming He, Georgia Gkioxari, Piotr Dollár, and Ross Girshick.
\newblock Mask r-cnn, 2018.

\bibitem[Hyeon et~al.(2025)Hyeon, Jung, Kim, Ryu, and Do]{hyeon2025mathreadertexttospeechmathematical}
Sieun Hyeon, Kyudan Jung, Nam-Joon Kim, Hyun~Gon Ryu, and Jaeyoung Do.
\newblock Mathreader : Text-to-speech for mathematical documents, 2025.

\bibitem[Landeghem et~al.(2023)Landeghem, Tito, Łukasz Borchmann, Pietruszka, Józiak, Powalski, Jurkiewicz, Coustaty, Ackaert, Valveny, Blaschko, Moens, and Stanisławek]{vanlandeghem2023documentunderstandingdatasetevaluation}
Jordy~Van Landeghem, Rubén Tito, Łukasz Borchmann, Michał Pietruszka, Paweł Józiak, Rafał Powalski, Dawid Jurkiewicz, Mickaël Coustaty, Bertrand Ackaert, Ernest Valveny, Matthew Blaschko, Sien Moens, and Tomasz Stanisławek.
\newblock Document understanding dataset and evaluation (dude), 2023.

\bibitem[Li et~al.(2016)Li, Galley, Brockett, Gao, and Dolan]{li2016diversitypromotingobjectivefunctionneural}
Jiwei Li, Michel Galley, Chris Brockett, Jianfeng Gao, and Bill Dolan.
\newblock A diversity-promoting objective function for neural conversation models, 2016.

\bibitem[Li et~al.(2022)Li, Xu, Lv, Cui, Zhang, and Wei]{li2022ditselfsupervisedpretrainingdocument}
Junlong Li, Yiheng Xu, Tengchao Lv, Lei Cui, Cha Zhang, and Furu Wei.
\newblock Dit: Self-supervised pre-training for document image transformer, 2022.

\bibitem[Li et~al.(2023)Li, Liu, Xiong, Yu, Gu, Liu, and Yu]{li2023structureawarelanguagemodelpretraining}
Xinze Li, Zhenghao Liu, Chenyan Xiong, Shi Yu, Yu Gu, Zhiyuan Liu, and Ge Yu.
\newblock Structure-aware language model pretraining improves dense retrieval on structured data, 2023.

\bibitem[Lin et~al.(2017)Lin, Dollár, Girshick, He, Hariharan, and Belongie]{lin2017featurepyramidnetworksobject}
Tsung-Yi Lin, Piotr Dollár, Ross Girshick, Kaiming He, Bharath Hariharan, and Serge Belongie.
\newblock Feature pyramid networks for object detection, 2017.

\bibitem[Liu et~al.(2021)Liu, Lin, Cao, Hu, Wei, Zhang, Lin, and Guo]{liu2021swintransformerhierarchicalvision}
Ze Liu, Yutong Lin, Yue Cao, Han Hu, Yixuan Wei, Zheng Zhang, Stephen Lin, and Baining Guo.
\newblock Swin transformer: Hierarchical vision transformer using shifted windows, 2021.

\bibitem[Loshchilov and Hutter(2017)]{loshchilov2017sgdrstochasticgradientdescent}
Ilya Loshchilov and Frank Hutter.
\newblock Sgdr: Stochastic gradient descent with warm restarts, 2017.

\bibitem[Loshchilov and Hutter(2019)]{loshchilov2019decoupledweightdecayregularization}
Ilya Loshchilov and Frank Hutter.
\newblock Decoupled weight decay regularization, 2019.

\bibitem[Ma et~al.(2023)Ma, Du, Hu, Zhang, Zhang, Zhu, and Liu]{ma2023hrdocdatasetbaselinemethod}
Jiefeng Ma, Jun Du, Pengfei Hu, Zhenrong Zhang, Jianshu Zhang, Huihui Zhu, and Cong Liu.
\newblock Hrdoc: Dataset and baseline method toward hierarchical reconstruction of document structures, 2023.

\bibitem[Pawlik and Augsten(2015)]{10.1145/2699485}
Mateusz Pawlik and Nikolaus Augsten.
\newblock Efficient computation of the tree edit distance.
\newblock \emph{ACM Trans. Database Syst.}, 40\penalty0 (1), 2015.

\bibitem[Pawlik and Augsten(2016)]{PAWLIK2016157}
Mateusz Pawlik and Nikolaus Augsten.
\newblock Tree edit distance: Robust and memory-efficient.
\newblock \emph{Information Systems}, 56:\penalty0 157--173, 2016.

\bibitem[Qiang et~al.(2019)Qiang, Fu, Yu, Guo, Zhou, and Sigal]{1547260908565-217496438}
Yu-Ting Qiang, Yan-Wei Fu, Xiao Yu, Yan-Wen Guo, Zhi-Hua Zhou, and Leonid Sigal.
\newblock {Learning to Generate Posters of Scientific Papers by Probabilistic Graphical Models}.
\newblock \emph{Journal of Computer Science and Technology}, 34\penalty0 (1):\penalty0 155--169, 2019.

\bibitem[Rausch et~al.(2021)Rausch, Martinez, Bissig, Zhang, and Feuerriegel]{rausch2021docparserhierarchicalstructureparsing}
Johannes Rausch, Octavio Martinez, Fabian Bissig, Ce Zhang, and Stefan Feuerriegel.
\newblock Docparser: Hierarchical structure parsing of document renderings, 2021.

\bibitem[Ren et~al.(2016)Ren, He, Girshick, and Sun]{ren2016fasterrcnnrealtimeobject}
Shaoqing Ren, Kaiming He, Ross Girshick, and Jian Sun.
\newblock Faster r-cnn: Towards real-time object detection with region proposal networks, 2016.

\bibitem[See et~al.(2017)See, Liu, and Manning]{see2017pointsummarizationpointergeneratornetworks}
Abigail See, Peter~J. Liu, and Christopher~D. Manning.
\newblock Get to the point: Summarization with pointer-generator networks, 2017.

\bibitem[Smith(2007)]{4376991}
R. Smith.
\newblock An overview of the tesseract ocr engine.
\newblock In \emph{Ninth International Conference on Document Analysis and Recognition (ICDAR 2007)}, pages 629--633, 2007.

\bibitem[Sun et~al.(2025)Sun, Pan, Yang, Sui, Shi, Cheng, Li, Huang, Zhang, Yang, and Li]{sun2025p2pautomatedpapertopostergeneration}
Tao Sun, Enhao Pan, Zhengkai Yang, Kaixin Sui, Jiajun Shi, Xianfu Cheng, Tongliang Li, Wenhao Huang, Ge Zhang, Jian Yang, and Zhoujun Li.
\newblock P2p: Automated paper-to-poster generation and fine-grained benchmark, 2025.

\bibitem[Tanaka et~al.(2024)Tanaka, Wang, and Ushiku]{Tanaka_2024_BMVC}
Shohei Tanaka, Hao Wang, and Yoshitaka Ushiku.
\newblock Scipostlayout: A dataset for layout analysis and layout generation of scientific posters.
\newblock In \emph{35th British Machine Vision Conference 2024, {BMVC} 2024, Glasgow, UK, November 25-28, 2024}. BMVA, 2024.

\bibitem[Tito et~al.(2023)Tito, Karatzas, and Valveny]{tito2023hierarchicalmultimodaltransformersmultipage}
Rubèn Tito, Dimosthenis Karatzas, and Ernest Valveny.
\newblock Hierarchical multimodal transformers for multi-page docvqa, 2023.

\bibitem[Vaswani et~al.(2023)Vaswani, Shazeer, Parmar, Uszkoreit, Jones, Gomez, Kaiser, and Polosukhin]{vaswani2023attentionneed}
Ashish Vaswani, Noam Shazeer, Niki Parmar, Jakob Uszkoreit, Llion Jones, Aidan~N. Gomez, Lukasz Kaiser, and Illia Polosukhin.
\newblock Attention is all you need, 2023.

\bibitem[Vinyals et~al.(2015)Vinyals, Toshev, Bengio, and Erhan]{vinyals2015tellneuralimagecaption}
Oriol Vinyals, Alexander Toshev, Samy Bengio, and Dumitru Erhan.
\newblock Show and tell: A neural image caption generator, 2015.

\bibitem[Wang et~al.(2024)Wang, Hu, Zhong, Sun, and Huo]{wang2024detectorderconstructtreeconstructionbased}
Jiawei Wang, Kai Hu, Zhuoyao Zhong, Lei Sun, and Qiang Huo.
\newblock Detect-order-construct: A tree construction based approach for hierarchical document structure analysis, 2024.

\bibitem[Wang et~al.(2025)Wang, Hu, and Huo]{wang2025unihdsaunifiedrelationprediction}
Jiawei Wang, Kai Hu, and Qiang Huo.
\newblock Unihdsa: A unified relation prediction approach for hierarchical document structure analysis, 2025.

\bibitem[Wang et~al.(2023)Wang, Dai, Chen, Huang, Li, Zhu, Hu, Lu, Lu, Li, Wang, and Qiao]{wang2023internimageexploringlargescalevision}
Wenhai Wang, Jifeng Dai, Zhe Chen, Zhenhang Huang, Zhiqi Li, Xizhou Zhu, Xiaowei Hu, Tong Lu, Lewei Lu, Hongsheng Li, Xiaogang Wang, and Yu Qiao.
\newblock Internimage: Exploring large-scale vision foundation models with deformable convolutions, 2023.

\bibitem[Wang et~al.(2021)Wang, Xu, Cui, Shang, and Wei]{wang2021layoutreaderpretrainingtextlayout}
Zilong Wang, Yiheng Xu, Lei Cui, Jingbo Shang, and Furu Wei.
\newblock Layoutreader: Pre-training of text and layout for reading order detection, 2021.

\bibitem[Xiao et~al.(2023)Xiao, Zhang, Wang, Tan, He, Zhao, Soong, and Lee]{Xiao2023}
Yujia Xiao, Shaofei Zhang, Xi Wang, Xu Tan, Lei He, Sheng Zhao, Frank~K. Soong, and Tan Lee.
\newblock Contextspeech: Expressive and efficient text-to-speech for paragraph reading.
\newblock In \emph{INTERSPEECH 2023}, page 4883–4887. ISCA, 2023.

\bibitem[Xing et~al.(2024)Xing, Cheng, Gao, Shao, Yu, Bu, Zheng, and Yao]{xing-etal-2024-dochienet}
Hangdi Xing, Changxu Cheng, Feiyu Gao, Zirui Shao, Zhi Yu, Jiajun Bu, Qi Zheng, and Cong Yao.
\newblock {D}oc{H}ie{N}et: A large and diverse dataset for document hierarchy parsing.
\newblock In \emph{Proceedings of the 2024 Conference on Empirical Methods in Natural Language Processing}, pages 1129--1142, Miami, Florida, USA, 2024. Association for Computational Linguistics.

\bibitem[Xu and Wan(2021)]{xu2021neural}
Sheng Xu and Xiaojun Wan.
\newblock {Neural Content Extraction for Poster Generation of Scientific Papers}, 2021.

\bibitem[Zhang et~al.(2023)Zhang, Guo, Tu, Chen, Tang, Zhu, Zhang, and Gui]{zhang2023readingordermattersinformation}
Chong Zhang, Ya Guo, Yi Tu, Huan Chen, Jinyang Tang, Huijia Zhu, Qi Zhang, and Tao Gui.
\newblock Reading order matters: Information extraction from visually-rich documents by token path prediction, 2023.

\bibitem[Zhang et~al.(2024)Zhang, Tu, Zhao, Yuan, Chen, Zhang, Chai, Guo, Zhu, Zhang, and Gui]{zhang-etal-2024-modeling}
Chong Zhang, Yi Tu, Yixi Zhao, Chenshu Yuan, Huan Chen, Yue Zhang, Mingxu Chai, Ya Guo, Huijia Zhu, Qi Zhang, and Tao Gui.
\newblock Modeling layout reading order as ordering relations for visually-rich document understanding.
\newblock In \emph{Proceedings of the 2024 Conference on Empirical Methods in Natural Language Processing}, pages 9658--9678, Miami, Florida, USA, 2024. Association for Computational Linguistics.

\end{thebibliography}
}


\clearpage
\clearpage
\appendix

\section{Dataset Description}

This section presents the annotation guideline and detailed statistics of the dataset.

\subsection{Annotation Guideline}

The following guideline defines the procedure for annotating SciPostLayout~\cite{Tanaka_2024_BMVC} posters as DFS-ordered trees.
It was developed through discussions between the in-house supervisors of the external vendor and the authors.

\begin{itemize}
    \item The Root node represents the poster.
    \item Each BBox is assigned one of eight categories: Title, Author Info, Section, Text, List, Table, Figure, or Caption.
    \item Title, Author Info, and Section are always assigned the Root as their parent.
    \item Text, List, Table, and Figure are assigned the Section that contains them, if such a Section exists; otherwise, their parent is the Root.
    \item A Caption is assigned its corresponding Figure or Table as its parent.
    \item When a node has multiple children, they are ordered according to their reading priority. The DFS traversal follows this order.
\end{itemize}



\subsection{Annotation Consistency}


This subsection describes the evaluation procedure for the annotation consistency reported in Section~\ref{sec:annotation}.
We evaluated agreement on 100 randomly selected posters from the test set by comparing the original annotations with independently annotated trees by two additional annotators.
Agreement was computed for the entire tree, parent–child relations, and reading-order relations.
Tree agreement was measured using STEDS defined in Section~\ref{sec:metrics}.
Parent–child agreement was defined as the proportion of nodes whose assigned parent matches between two annotations.
Reading-order agreement was defined as the proportion of consecutive node pairs in the DFS traversal that match between two annotations.
Each score was computed per poster and averaged over the 100 posters.
As a result, STEDS = 0.91, parent–child pair agreement = 0.97, and reading-order pair agreement = 0.94, indicating high agreement.

\begin{figure}[t!]
\centering
\includegraphics[width=\columnwidth]{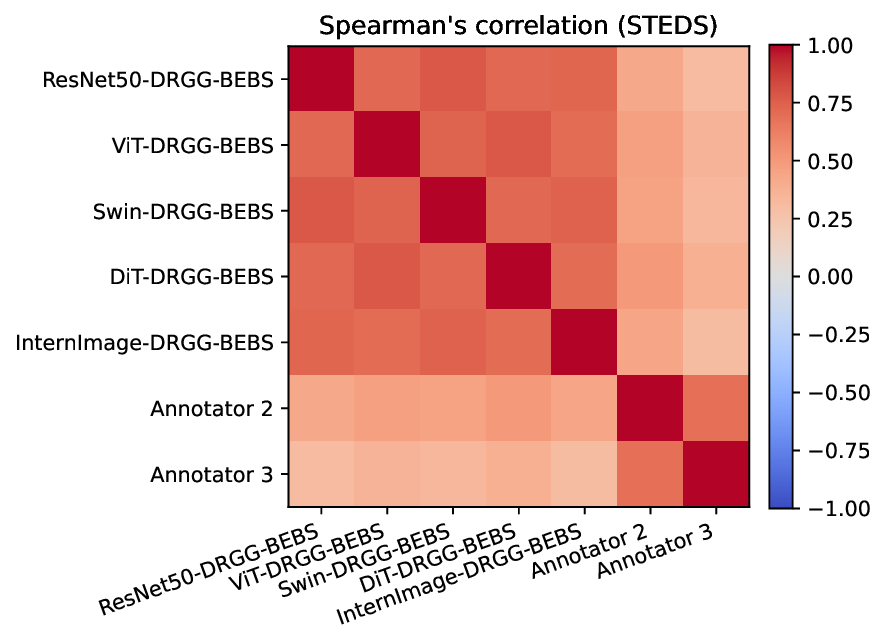}
\caption{
Heatmap of Spearman’s rank correlation coefficients between the STEDS of DRGG-BEBS models and those of the annotators.
}
\label{fig:spearman_heatmap}
\end{figure}

In addition, both the model STEDS in the evaluation experiments and the inter-annotator STEDS are approximately 0.9, raising the possibility that model errors may stem from inconsistent annotations.
To investigate this hypothesis, we computed the Spearman rank correlation between per-poster model STEDS and inter-annotator STEDS.
The average correlation is moderate (Spearman’s $\rho$ = 0.40), indicating that model errors are not primarily attributable to inconsistent annotations.
The full correlation matrix is shown in Figure~\ref{fig:spearman_heatmap}.
As shown in the figure, correlations among models and among annotators are high, whereas correlations between models and annotators are low.

\subsection{Detailed Statistics of Dataset}

This subsection provides detailed statistics that supplement the figures and tables presented in the main paper.

\subsubsection{Layout Tree Statistics}

Table~\ref{tab:tree_statistics} summarizes statistics of the tree structures in SciPostLayoutTree and DocHieNet~\cite{xing-etal-2024-dochienet}.
This result is consistent with the conclusion drawn from Fig.~\ref{fig:tree_stats_summary}.

\subsubsection{Reading Order Statistics}

Tables~\ref{tab:ro_direction_distance}--\ref{tab:dhn_ro_direction_distance} provide the underlying counts for the heatmaps shown in Fig.~\ref{fig:ro_rose_heatmap}.

\begin{figure}[t!]
\centering
\includegraphics[width=\columnwidth]{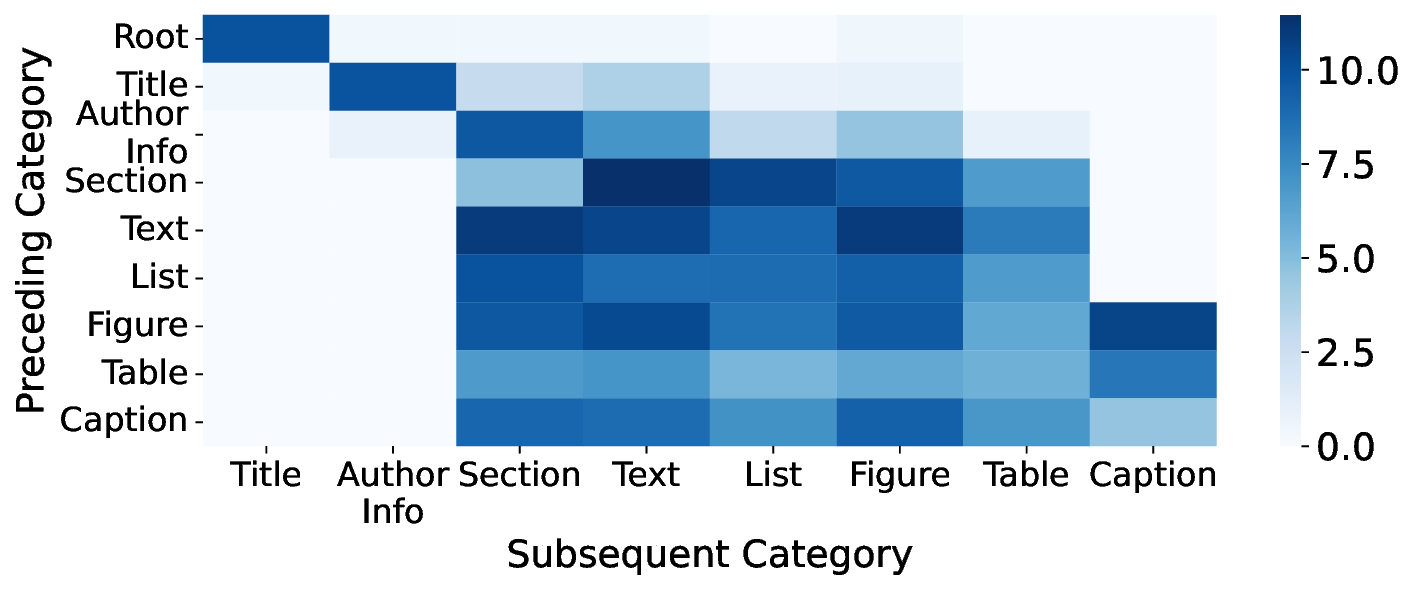}
\caption{
Heatmap of reading order frequencies in SciPostLayoutTree, aggregated by transitions between categories, with counts normalized per 1,000 pages.
Heatmap values represent $\log_2(1 + \text{count})$.
}
\label{fig:ro_category_heatmap}
\end{figure}

The category transition patterns in Figure~\ref{fig:ro_category_heatmap} and Table~\ref{tab:ro_category_frequency} indicate a consistent introductory sequence (Root~$\rightarrow$~Title~$\rightarrow$~Author Info), followed by diverse transitions such as Section~$\rightarrow$~Text and Text~$\rightarrow$~Figure.


\begin{table}[t!]
\centering
\small
\caption{
Statistics of the tree structures.
Tree depth denotes the maximum number of nodes from the root to any leaf.
Tree width denotes the maximum number of nodes at any single depth level.
Children per node denotes the average number of child nodes per parent node.
}
\resizebox{\columnwidth}{!}{
\begin{tabular}{l|rrr}
\toprule
Dataset & Tree Depth & Tree Width & Children per Node \\
\midrule
SciPostLayoutTree & 3.37\ ($\pm$ 0.56) & 15.24\ ($\pm$ 7.49) & 0.96\ ($\pm$ 2.47) \\
DocHieNet & 3.16\ ($\pm$ 0.90) & 9.41\ ($\pm$ 6.60) & 0.93\ ($\pm$ 2.78) \\
\bottomrule
\end{tabular}
}
\label{tab:tree_statistics}
\end{table}

\begin{figure}[t!]
\centering
\includegraphics[width=\columnwidth]{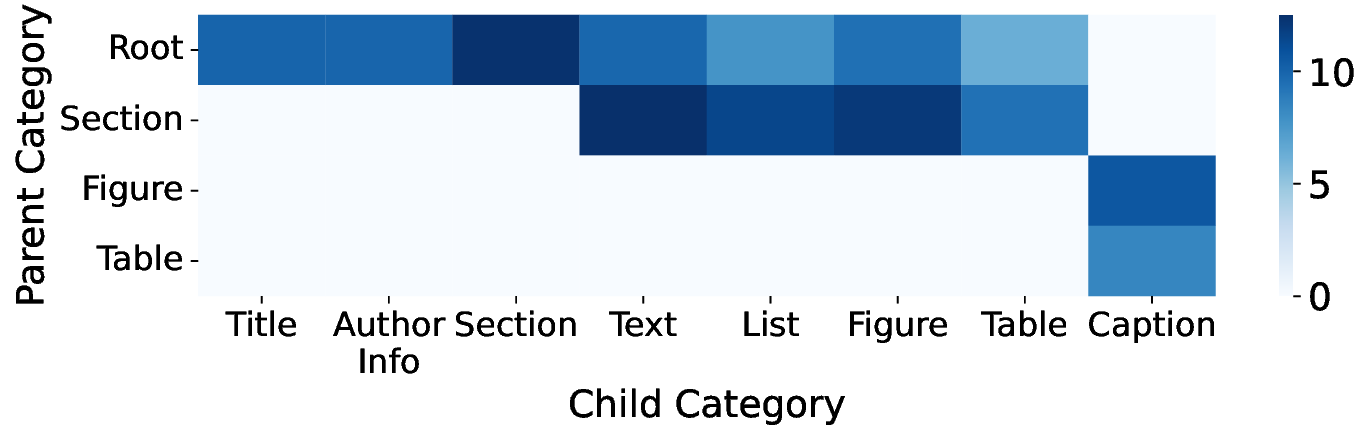}
\caption{
Heatmap of parent-child relation frequencies in SciPostLayoutTree, aggregated by transitions between categories.
The format follows Fig.~\ref{fig:ro_category_heatmap}.
}
\label{fig:pc_category_heatmap}
\end{figure}

\subsubsection{Parent-Child Relation Statistics}

Tables~\ref{tab:pc_direction_distance}--\ref{tab:dhn_pc_direction_distance} provide the underlying counts for the heatmaps shown in Fig.~\ref{fig:pc_rose_heatmap}.

Figure~\ref{fig:pc_category_heatmap} and Table~\ref{tab:pc_category_frequency} reveal that the category transition patterns follow the annotation guideline, for example in transitions such as Section~$\rightarrow$~Text and Figure~$\rightarrow$~Caption, indicating that parent-child relations depend on the categories.

\begin{table*}[t!]
\centering
\small
\caption{
Reading order frequency in SciPostLayoutTree, aggregated by direction and normalized distance bins, with counts normalized per 1,000 pages.
}
\begin{tabular}{l|rrrrrr}
\toprule
Direction \textbackslash{} Distance & (0, 1] & (1, 2] & (2, 4] & (4, 8] & (8, 16] & (16, $\infty$) \\
\midrule
Right         & 2,143.97 & 889.16 & 41.15 & 5.48 & 0.38 & 0.00 \\
Bottom-Right  & 2,045.61 & 150.72 & 23.57 & 7.64 & 4.46 & 0.76 \\
Bottom        & 7,800.48 & 1,913.36 & 200.03 & 78.74 & 48.16 & 22.93 \\
Bottom-Left   & 2,602.24 & 483.50 & 78.99 & 22.81 & 13.38 & 5.35 \\
Left          & 1,568.61 & 524.53 & 247.29 & 57.20 & 4.59 & 0.38 \\
Top-Left      & 130.97 & 14.27 & 8.28 & 7.01 & 4.20 & 1.15 \\
Top           & 390.50 & 35.42 & 72.37 & 120.52 & 78.74 & 96.70 \\
Top-Right     & 390.24 & 514.33 & 261.18 & 190.60 & 117.08 & 117.47 \\
\bottomrule
\end{tabular}
\label{tab:ro_direction_distance}
\end{table*}

\begin{table*}[t!]
\centering
\small
\caption{
Reading order frequency in DocHieNet, aggregated by direction and normalized distance bins, with counts normalized per 1,000 pages.
}
\begin{tabular}{l|rrrrrr}
\toprule
Direction \textbackslash{} Distance & (0, 1] & (1, 2] & (2, 4] & (4, 8] & (8, 16] & (16, $\infty$) \\
\midrule
Right         & 1,358.60 & 412.46 & 127.70 & 49.80 & 17.91 & 3.20 \\
Bottom-Right  & 1,207.44 & 219.20 & 50.32 & 13.01 & 9.87 & 7.97 \\
Bottom        & 3,322.14 & 2,036.66 & 192.67 & 58.62 & 43.13 & 52.41 \\
Bottom-Left   & 1,177.77 & 246.45 & 57.71 & 20.06 & 12.42 & 11.11 \\
Left          & 1,136.85 & 398.86 & 128.10 & 35.49 & 18.10 & 5.36 \\
Top-Left      & 10.98 & 28.56 & 11.50 & 7.06 & 5.82 & 4.31 \\
Top           & 35.88 & 11.37 & 19.02 & 26.40 & 26.53 & 27.45 \\
Top-Right     & 30.39 & 63.59 & 34.31 & 32.81 & 15.69 & 7.91 \\
\bottomrule
\end{tabular}
\label{tab:dhn_ro_direction_distance}
\end{table*}

\begin{table*}[t!]
\centering
\small
\caption{
Reading order frequency in SciPostLayoutTree, aggregated by transitions between categories, with counts normalized per 1,000 pages.
}
\begin{tabular}{l|rrrrrrrr}
\toprule
Preceding \textbackslash{} Subsequent & Title & Author
Info & Section & Text & List & Figure & Table & Caption \\
\midrule
Root & 998.85 & 0.25 & 0.25 & 0.25 & 0.00 & 0.38 & 0.00 & 0.00 \\
Title & 0.25 & 978.98 & 6.24 & 11.98 & 0.76 & 0.89 & 0.00 & 0.00 \\
Author
Info & 0.00 & 0.76 & 820.61 & 125.62 & 7.39 & 23.19 & 0.89 & 0.00 \\
Section & 0.00 & 0.00 & 26.88 & 2,813.73 & 1,498.53 & 802.78 & 101.54 & 0.00 \\
Text & 0.00 & 0.00 & 1,946.24 & 1,491.02 & 538.03 & 1,991.97 & 283.48 & 0.00 \\
List & 0.00 & 0.00 & 1,000.89 & 427.83 & 439.55 & 646.83 & 101.67 & 0.00 \\
Figure & 0.00 & 0.00 & 826.22 & 1,274.30 & 364.12 & 766.85 & 65.23 & 1,555.61 \\
Table & 0.00 & 0.00 & 106.77 & 128.55 & 38.35 & 64.85 & 47.39 & 327.81 \\
Caption & 0.00 & 0.00 & 530.39 & 444.13 & 140.53 & 623.26 & 120.27 & 23.32 \\
\bottomrule
\end{tabular}
\label{tab:ro_category_frequency}
\end{table*}

\begin{table*}[t!]
\centering
\small
\caption{
Parent-child relation frequency in SciPostLayoutTree, aggregated by direction and normalized distance bins, with counts normalized per 1,000 pages.
}
\begin{tabular}{l|rrrrrr}
\toprule
Direction \textbackslash{} Distance & (0, 1] & (1, 2] & (2, 4] & (4, 8] & (8, 16] & (16, $\infty$) \\
\midrule
Right         & 1,173.53 & 568.73 & 312.01 & 89.69 & 10.19 & 0.51 \\
Bottom-Right  & 2,256.08 & 734.74 & 501.08 & 269.08 & 121.29 & 57.84 \\
Bottom        & 3,367.94 & 1,016.94 & 1,144.86 & 851.83 & 495.35 & 329.72 \\
Bottom-Left   & 565.17 & 199.77 & 130.34 & 70.96 & 50.07 & 25.35 \\
Left          & 159.38 & 67.40 & 16.94 & 3.19 & 0.00 & 0.00 \\
Top-Left      & 114.79 & 3.95 & 1.27 & 1.15 & 0.00 & 0.00 \\
Top           & 369.35 & 6.24 & 2.42 & 4.71 & 2.42 & 2.04 \\
Top-Right     & 42.81 & 140.02 & 85.36 & 41.92 & 15.93 & 15.03 \\
\bottomrule
\end{tabular}
\label{tab:pc_direction_distance}
\end{table*}

\begin{table*}[t!]
\centering
\small
\caption{
Parent-child relation frequency in DocHieNet, aggregated by direction and normalized distance bins, with counts normalized per 1,000 pages.
}
\begin{tabular}{l|rrrrrr}
\toprule
Direction \textbackslash{} Distance & (0, 1] & (1, 2] & (2, 4] & (4, 8] & (8, 16] & (16, $\infty$) \\
\midrule
Right         & 764.85 & 211.82 & 27.78 & 3.07 & 0.13 & 0.00 \\
Bottom-Right  & 1,054.83 & 223.65 & 223.97 & 260.83 & 199.53 & 85.62 \\
Bottom        & 523.82 & 365.73 & 471.28 & 564.41 & 549.51 & 480.10 \\
Bottom-Left   & 219.01 & 58.04 & 63.66 & 80.58 & 68.88 & 42.28 \\
Left          & 118.95 & 16.08 & 3.66 & 0.07 & 0.00 & 0.00 \\
Top-Left      & 3.86 & 0.46 & 0.33 & 0.07 & 0.00 & 0.00 \\
Top           & 25.95 & 2.29 & 1.83 & 1.57 & 1.50 & 3.01 \\
Top-Right     & 9.74 & 104.76 & 30.06 & 27.25 & 31.24 & 32.09 \\
\bottomrule
\end{tabular}
\label{tab:dhn_pc_direction_distance}
\end{table*}

\begin{table*}[t!]
\centering
\small
\caption{
Parent-child relaiton frequency in SciPostLayoutTree, aggregated by transitions between categories, with counts normalized per 1,000 pages.
}
\begin{tabular}{l|rrrrrrrr}
\toprule
\multicolumn{1}{l}{Parent \textbackslash{} Child} & \multicolumn{1}{|l}{Title} & \multicolumn{1}{l}{Author
Info} & \multicolumn{1}{l}{Section} & \multicolumn{1}{l}{Text} & \multicolumn{1}{l}{List} & \multicolumn{1}{l}{Figure} & \multicolumn{1}{l}{Table} & \multicolumn{1}{l}{Caption} \\
\midrule
Root & 999.11 & 980.00 & 5,264.49 & 902.41 & 206.14 & 669.51 & 75.42 & 0.00 \\
Section & 0.00 & 0.00 & 0.00 & 5,815.01 & 2,821.12 & 4,251.50 & 645.05 & 0.00 \\
Figure & 0.00 & 0.00 & 0.00 & 0.00 & 0.00 & 0.00 & 0.00 & 1,577.27 \\
Table & 0.00 & 0.00 & 0.00 & 0.00 & 0.00 & 0.00 & 0.00 & 329.47 \\
\bottomrule
\end{tabular}
\label{tab:pc_category_frequency}
\end{table*}

\clearpage
\clearpage
\section{Experimental Details}

This section provides the implementation details and the experimental setup.

\subsection{Implementation Details}

We conducted all experiments using publicly available pretrained visual backbones, including ResNet-50~\cite{he2015deepresiduallearningimage}, ViT-Base~\cite{dosovitskiy2021imageworth16x16words}, Swin-Base~\cite{liu2021swintransformerhierarchicalvision}, DiT-Base~\cite{li2022ditselfsupervisedpretrainingdocument}, and InternImage-Base~\cite{wang2023internimageexploringlargescalevision}.
All backbone parameters were fine-tuned during training.
All model parameters, except those in the pretrained backbones, were initialized using PyTorch's default initialization schemes.
Random seeds were not explicitly fixed, and deterministic computation settings were not enforced; therefore, results may exhibit minor variations across runs due to stochastic training dynamics.

We employed the AdamW optimizer~\cite{loshchilov2019decoupledweightdecayregularization} with an initial learning rate of $2 \times 10^{-4}$, $\beta_1 = 0.9$, $\beta_2 = 0.999$, and a weight decay of 0.05. A uniform learning rate scaling factor of 1.0 was applied to all layers.
The learning rate was scheduled using a linear warmup~\cite{vaswani2023attentionneed} over the first 500 iterations with a warmup factor of 0.01, followed by cosine decay~\cite{loshchilov2017sgdrstochasticgradientdescent} to zero over the remaining 10,500 iterations, totaling 11,000 training steps (approximately 24 epochs).
Gradient clipping with a maximum norm of 1.0 and $L^2$-norm type was applied to stabilize training.

Training was performed on 8 GPUs with a total batch size of 16, corresponding to 2 samples per GPU.
All hyperparameters were selected based on performance on the validation set.
Each model was evaluated on the test set in a single run with the final model checkpoint.
Statistical significance testing was performed to assess performance differences.

\subsection{Experimental Setup}

All experiments were conducted on a computing node equipped with 8 NVIDIA A100-SXM4-80GB GPUs (total 640 GB GPU memory) and 2 AMD EPYC 7713 64-core processors, resulting in 128 physical CPU cores.
The system had 2.0 TiB of RAM, of which approximately 1.9 TiB was available.
The operating system was Ubuntu 22.04.5 LTS (Jammy Jellyfish).
The GPU driver version was 535.161.08 with CUDA 12.2, while PyTorch was built with CUDA 12.1 support.
The software environment consisted of Python 3.10.10, PyTorch 2.5.1+cu121, transformers 4.52.3, detectron2 0.6, and detrex 0.3.0.

\clearpage
\clearpage
\section{Additional Experiments}

This section presents additional figures, tables, and experiments that supplement the analysis in the main paper.

\subsection{Effect of Text Embedding}

We explored a model that incorporates textual content in posters as additional features.
For each BBox $b_i$ other than the Figure BBoxes, OCR text $t_i$ was extracted using Tesseract~\cite{4376991}.
Each $t_i$ was then encoded into a text feature vector $\mathbf{u_i}$ using a pretrained SciBERT model~\cite{beltagy2019scibertpretrainedlanguagemodel}, where $\mathbf{u_i}$ denotes the mean-pooled output of the last layer.
For the Figure BBoxes, the special token \texttt{[FIGURE]} was fed into SciBERT to obtain corresponding text features. 
The text feature of the Root node, $\mathbf{u_0}$, was set to the zero vector.  
Finally, each text feature $\mathbf{u_i}$ was concatenated with the corresponding visual feature $\mathbf{v^r_i}$ or $\mathbf{v^c_i}$, as well as the BBox features $\mathbf{z_i}$ and $\mathbf{e_i}$, to obtain the multimodal features as follows:
\[
\mathbf{m^r_i} = \text{concat}(\mathbf{v^r_i}, \mathbf{z_i}, \mathbf{e_i}, \mathbf{u_i}),\ 
\mathbf{m^c_i} = \text{concat}(\mathbf{v^c_i}, \mathbf{z_i}, \mathbf{e_i}, \mathbf{u_i})
\]
The subsequent procedure is identical to that in the main paper.

We compare the models evaluated in the main paper with additional variants incorporating the text features.

\noindent
\textbf{DRGG-TE (Text Embedding)}
augments the input features with the text features.

\noindent
\textbf{DRGG-TEBS (Text Embedding and Beam Search)}
integrates DRGG-TE and DRGG-BS.

\noindent
\textbf{DRGG-BETE (BBox Embedding and Text Embedding)}
integrates DRGG-BE and DRGG-TE.

\noindent
\textbf{DRGG-BETEBS (BBox Embedding, Text Embedding, and Beam Search)}
integrates DRGG-BE, DRGG-TE, and DRGG-BS.

Table~\ref{tab:main_results_te} shows that models incorporating text features perform comparably to, or slightly worse than, those using only visual and BBox features.  
This observation may be attributed to the following factors.
First, humans typically interpret the layout structure of a poster prior to reading its textual content.
Analogously, even when the language is unreadable, humans can often infer the overall layout structure if basic properties such as reading direction (e.g., left-to-right) are known.
These considerations suggest that the contribution of textual content to structural analysis for posters is limited.
Second, as shown in Table~1, scientific posters often contain many figures, which reduces the importance of textual content.
Third, the current models predict reading order and parent-child relations using greedy or beam search, which are inherently limited to local decisions based on one or two decoding steps.  
Discourse-level structural understanding based on textual content requires a lookahead decoding strategy that can capture longer dependencies.

\begin{table}[th!]
\centering
\small
\caption{
Comparison of DRGG, its extensions, and their text-feature-enhanced variants across visual backbones.
${\ast}$ and ${\ast\ast}$ indicate significant improvements over DRGG at $p < 0.005$ and $p < 0.05$, respectively, according to Wilcoxon signed-rank test.
}
\setlength{\tabcolsep}{2pt}
\begin{tabular}{ll|rrr}
\toprule
Backbone & Decoder & STEDS ($\uparrow$) & REDS ($\uparrow$) & TED ($\downarrow$) \\
\midrule
\multirow{8}{*}{ResNet-50} & DRGG & 68.74 & 75.07 & 8.83 \\
    & DRGG-BE & 84.24$^{\ast}$ & 86.44$^{\ast}$ & 4.41$^{\ast}$ \\
    & DRGG-BS & 76.79$^{\ast}$ & 83.14$^{\ast}$ & 6.65$^{\ast}$ \\
    & DRGG-BEBS & 88.45$^{\ast}$ & 90.40$^{\ast}$ & 3.22$^{\ast}$ \\
    & DRGG-TE & 71.21$^{\ast}$ & 77.23$^{\ast}$ & 8.21$^{\ast}$ \\
    & DRGG-TEBS & 78.46$^{\ast}$ & 84.19$^{\ast}$ & 6.15$^{\ast}$ \\
    & DRGG-BETE & 82.66$^{\ast}$ & 85.08$^{\ast}$ & 4.90$^{\ast}$ \\
    & DRGG-BETEBS & 87.15$^{\ast}$ & 89.20$^{\ast}$ & 3.59$^{\ast}$ \\
\midrule
\multirow{8}{*}{ViT} & DRGG & 80.40 & 85.94 & 5.45 \\
    & DRGG-BE & 86.38$^{\ast}$ & 88.24$^{\ast}$ & 3.85$^{\ast}$ \\
    & DRGG-BS & 83.89$^{\ast}$ & 89.46$^{\ast}$ & 4.46$^{\ast}$ \\
    & DRGG-BEBS & 90.04$^{\ast}$ & 91.73$^{\ast}$ & 2.78$^{\ast}$ \\
    & DRGG-TE & 80.32 & 86.50 & 5.56 \\
    & DRGG-TEBS & 83.35$^{\ast}$ & 89.33$^{\ast}$ & 4.73$^{\ast}$ \\
    & DRGG-BETE & 85.74$^{\ast}$ & 87.82$^{\ast}$ & 4.00$^{\ast}$ \\
    & DRGG-BETEBS & 89.04$^{\ast}$ & 90.93$^{\ast}$ & 3.06$^{\ast}$ \\
\midrule
\multirow{8}{*}{Swin} & DRGG & 79.90 & 85.77 & 5.69 \\
    & DRGG-BE & 86.73$^{\ast}$ & 88.42$^{\ast}$ & 3.69$^{\ast}$ \\
    & DRGG-BS & 83.11$^{\ast}$ & 89.02$^{\ast}$ & 4.74$^{\ast}$ \\
    & DRGG-BEBS & 89.26$^{\ast}$ & 90.88$^{\ast}$ & 2.95$^{\ast}$ \\
    & DRGG-TE & 80.31 & 85.77 & 5.52 \\
    & DRGG-TEBS & 84.18$^{\ast}$ & 89.51$^{\ast}$ & 4.46$^{\ast}$ \\
    & DRGG-BETE & 84.72$^{\ast}$ & 86.51 & 4.22$^{\ast}$ \\
    & DRGG-BETEBS & 88.37$^{\ast}$ & 90.11$^{\ast}$ & 3.23$^{\ast}$ \\
\midrule
\multirow{8}{*}{DiT} & DRGG & 78.33 & 84.81 & 6.09 \\
    & DRGG-BE & 85.32$^{\ast}$ & 87.36$^{\ast}$ & 4.22$^{\ast}$ \\
    & DRGG-BS & 82.12$^{\ast}$ & 88.72$^{\ast}$ & 5.07$^{\ast}$ \\
    & DRGG-BEBS & 88.69$^{\ast}$ & 90.63$^{\ast}$ & 3.24$^{\ast}$ \\
    & DRGG-TE & 79.25 & 85.52 & 5.83 \\
    & DRGG-TEBS & 82.02$^{\ast}$ & 88.40$^{\ast}$ & 5.13$^{\ast}$ \\
    & DRGG-BETE & 85.56$^{\ast}$ & 87.71$^{\ast}$ & 4.10$^{\ast}$ \\
    & DRGG-BETEBS & 88.81$^{\ast}$ & 90.82$^{\ast}$ & 3.16$^{\ast}$ \\
\midrule
\multirow{8}{*}{InternImage} & DRGG & 80.19 & 85.53 & 5.53 \\
    & DRGG-BE & 86.89$^{\ast}$ & 88.56$^{\ast}$ & 3.70$^{\ast}$ \\
    & DRGG-BS & 83.75$^{\ast}$ & 89.25$^{\ast}$ & 4.55$^{\ast}$ \\
    & DRGG-BEBS & 89.34$^{\ast}$ & 90.97$^{\ast}$ & 2.93$^{\ast}$ \\
    & DRGG-TE & 81.62$^{\ast\ast}$ & 86.74 & 5.15 \\
    & DRGG-TEBS & 85.05$^{\ast}$ & 90.02$^{\ast}$ & 4.21$^{\ast}$ \\
    & DRGG-BETE & 84.69$^{\ast}$ & 86.63 & 4.26$^{\ast}$ \\
    & DRGG-BETEBS & 88.19$^{\ast}$ & 89.94$^{\ast}$ & 3.24$^{\ast}$ \\
\bottomrule
\end{tabular}
\label{tab:main_results_te}
\end{table}

\subsection{VLM Results}

\begin{figure}[t!]
\centering
\includegraphics[width=\columnwidth]{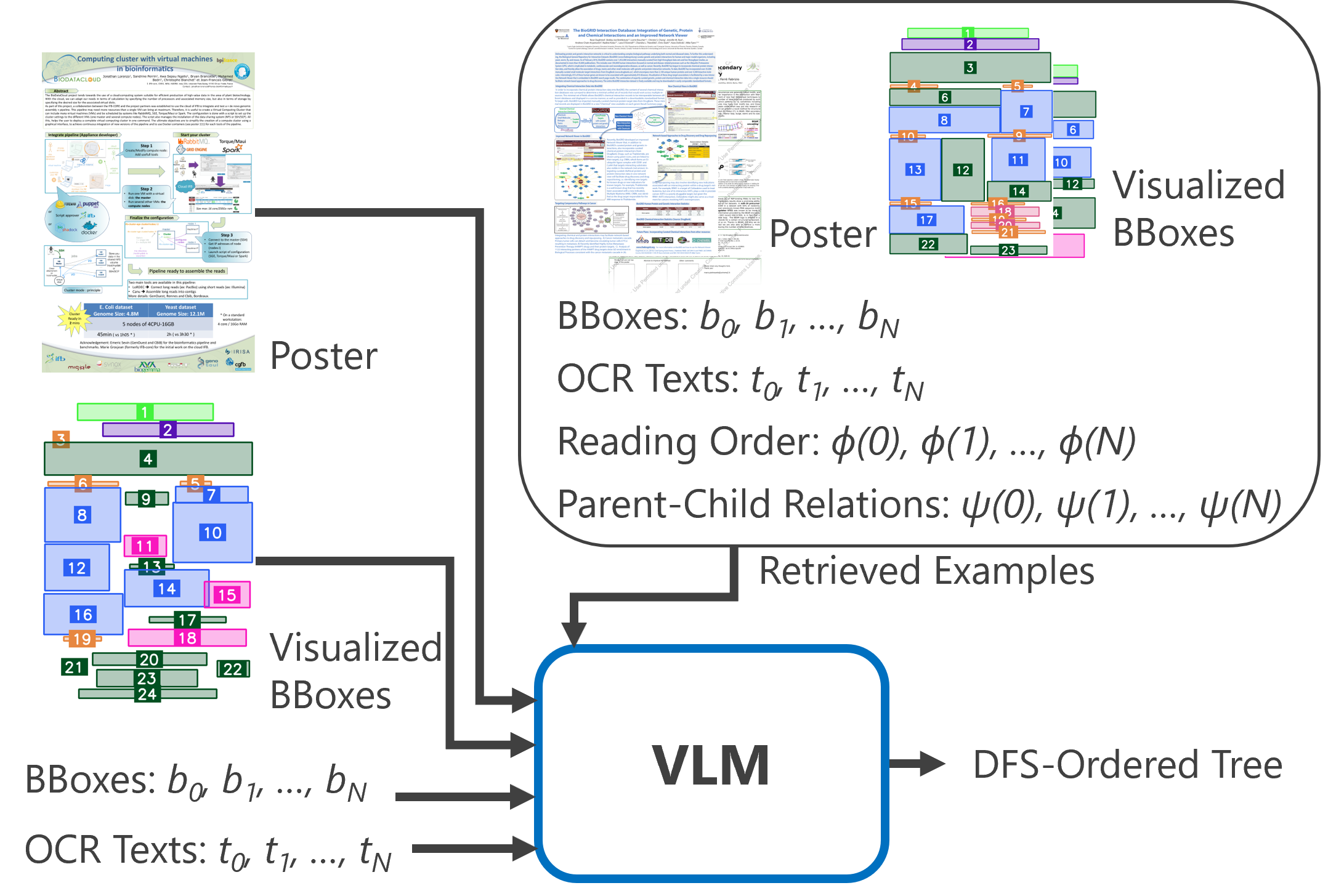}
\caption{
Overview of PromptTree.
}
\label{fig:promttree}
\end{figure}

We explored PromptTree, a model for decoding DFS-ordered trees using the latest VLMs such as GPT-5\footnote{https://openai.com/gpt-5/} and Gemini 3 Pro\footnote{https://deepmind.google/models/gemini/pro/}.
Figure~\ref{fig:promttree} provides an overview of PromptTree.
The model takes as input the following components: the poster image; a visualization of the BBoxes rendered on a white canvas of the same size as the poster, where BBoxes are color-coded by category and assigned indices after sorting in y–x order; the metadata of each BBox (including position, category, and BBox index); and OCR-extracted text.
In addition, two retrieved examples are provided as supplementary inputs.
These examples are selected from the training set by ranking candidate examples in descending order of Intersection over Union (IoU), which is computed via Hungarian algorithm between the BBoxes of the target image and those of each candidate.
Each retrieved example contains the same input information as the target image, along with the corresponding ground-truth reading order and parent–child relations.
The prompt template provided to the VLM is shown in Prompt~\ref{prompt:prompttree}.

\begin{table}[t!]
\centering
\small
\caption{
Comparison between DRGG variants and PromptTree, a VLM-based model.
}
\setlength{\tabcolsep}{2pt}
\begin{tabular}{ll|rrr}
\toprule
Backbone & Decoder & STEDS ($\uparrow$) & REDS ($\uparrow$) & TED ($\downarrow$) \\
\midrule
\multirow{2}{*}{ResNet-50} & DRGG & 68.74 & 75.07 & 8.83 \\
    & DRGG-BEBS & 88.45 & 90.40 & 3.22 \\
\midrule
\multirow{2}{*}{ViT} & DRGG & 80.40 & 85.94 & 5.45 \\
    & DRGG-BEBS & 90.04 & 91.73 & 2.78 \\
\midrule
\multirow{2}{*}{Swin} & DRGG & 79.90 & 85.77 & 5.69 \\
    & DRGG-BEBS & 89.26 & 90.88 & 2.95 \\
\midrule
\multirow{2}{*}{DiT} & DRGG & 78.33 & 84.81 & 6.09 \\
    & DRGG-BEBS & 88.69 & 90.63 & 3.24 \\
\midrule
\multirow{2}{*}{InternImage} & DRGG & 80.19 & 85.53 & 5.53 \\
    & DRGG-BEBS & 89.34 & 90.97 & 2.93 \\
\midrule
\multicolumn{2}{c|}{GPT-5} & 76.96 & 83.13 & 6.34 \\
\multicolumn{2}{c|}{Gemini 3 Pro} & 81.96 & 89.66 & 5.01 \\
\bottomrule
\end{tabular}
\label{tab:vlm_results}
\end{table}

Table~\ref{tab:vlm_results} presents a comparison between DRGG variants and PromptTree.
PromptTree shows lower performance than all DRGG-BEBS variants when evaluated with either GPT-5 or Gemini 3 Pro, indicating that poster structure analysis remains challenging for the latest VLMs.

\begin{figure*}[t!]
\begin{tcolorbox}
\small
\begin{verbatim}
You are an expert in document structure analysis.
You are given bounding boxes from a scientific poster and one or more images of
the poster.
Your task is to infer:
1) the reading order of the bboxes,
2) the parent-child relationships forming a rooted tree.

Example 1:

Input bboxes:
- bbox_number=<bbox_id>, category=<category_name>, x=<x_norm>, y=<y_norm>, 
w=<w_norm>, h=<h_norm>, text="<ocr_text_if_any>"
- ...

Images for Example 1:
[POSTER_IMAGE]
[BBOXES_IMAGE]

Output:
reading_order = <reading_order_list>
tree = <tree_list_of_{bbox_number,parent}_objects>

Example 2:

Input bboxes:
- bbox_number=<bbox_id>, category=<category_name>, x=<x_norm>, y=<y_norm>,
w=<w_norm>, h=<h_norm>, text="<ocr_text_if_any>"
- ...

Images for Example 2:
[POSTER_IMAGE]
[BBOXES_IMAGE]

Output:
reading_order = <reading_order_list>
tree = <tree_list_of_{bbox_number,parent}_objects>

Now solve the following example.

Input bboxes:
- bbox_number=<bbox_id>, category=<category_name>, x=<x_norm>, y=<y_norm>,
w=<w_norm>, h=<h_norm>, text="<ocr_text_if_any>"
- ...

Use both the textual description of the bboxes and the visual layout in the
images.
Output the result in the same format as the previous examples, using JSON only.
Do not include any explanations.

Images for the target example:
[POSTER_IMAGE]
[BBOXES_IMAGE]

\end{verbatim}
\normalsize
\end{tcolorbox}
\captionof{prompt}{Prompt used for decoding a DFS-ordered tree with PromptTree.}
\label{prompt:prompttree}
\end{figure*}

\subsection{Additional Challenging Examples}

\begin{figure*}[t!]
\centering
\includegraphics[width=\linewidth,height=\textheight,keepaspectratio]{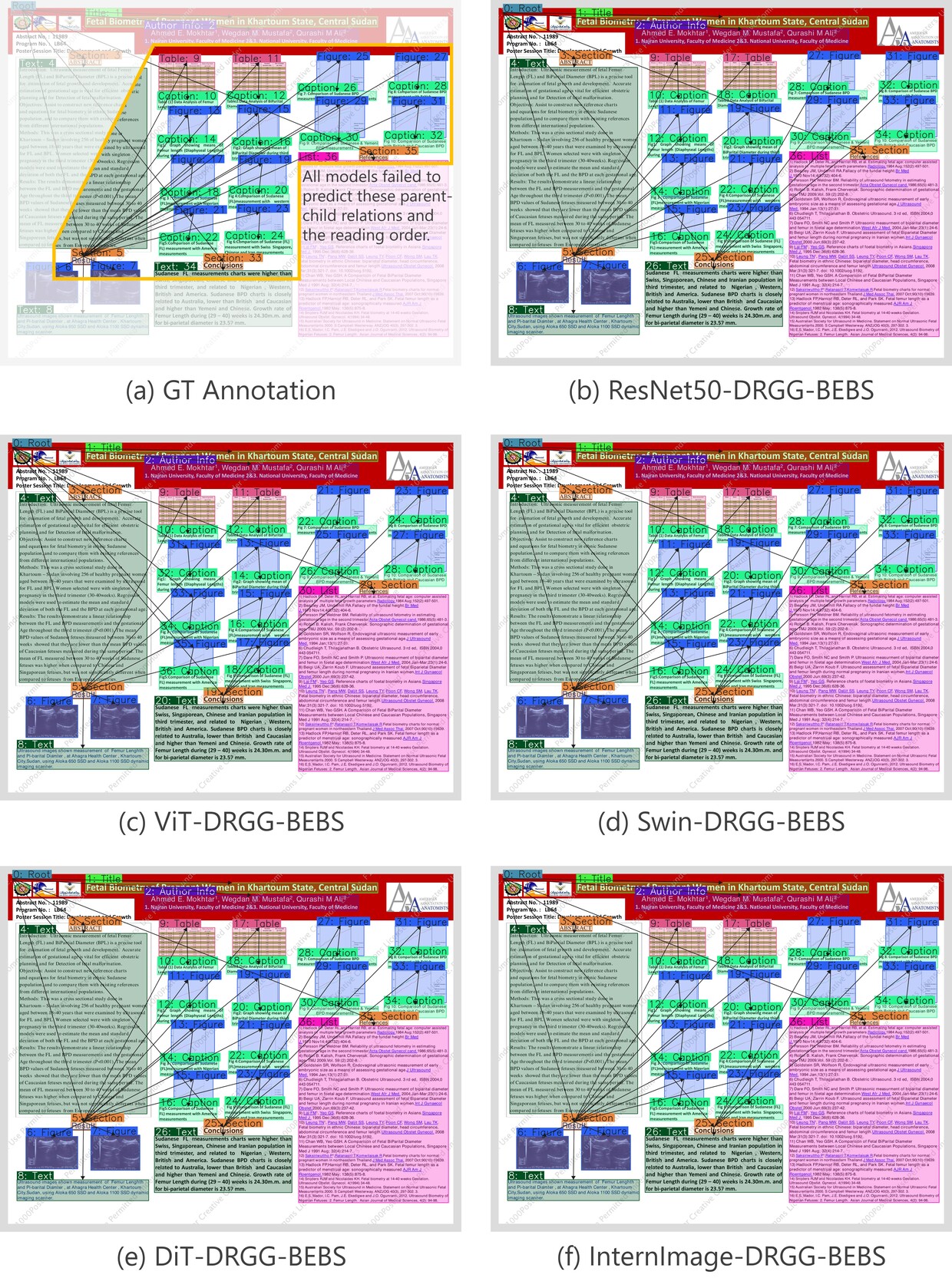}
\caption{
Example of a poster with the GT annotation and the predicted trees.
The predicted trees received low STEDS (42.70).
}
\label{fig:error_121663_all}
\end{figure*}

\begin{figure*}[t!]
\centering
\includegraphics[width=\linewidth,height=\textheight,keepaspectratio]{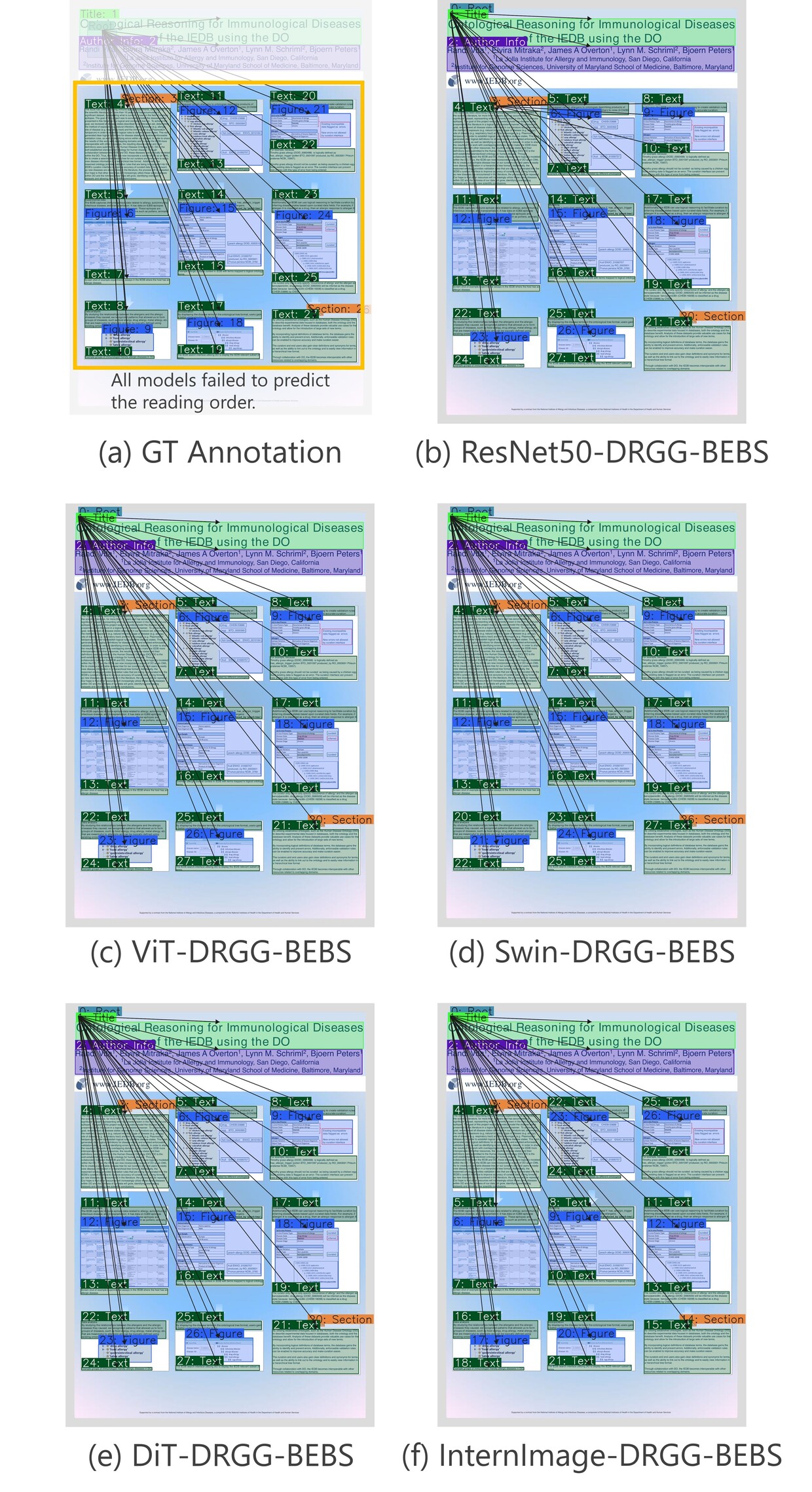}
\caption{
Example of a poster with the GT annotation and the predicted trees.
The predicted trees received low STEDS (25.71).
}
\label{fig:error_8858_all}
\end{figure*}

\begin{figure*}[t!]
\centering
\includegraphics[width=\linewidth,height=\textheight,keepaspectratio]{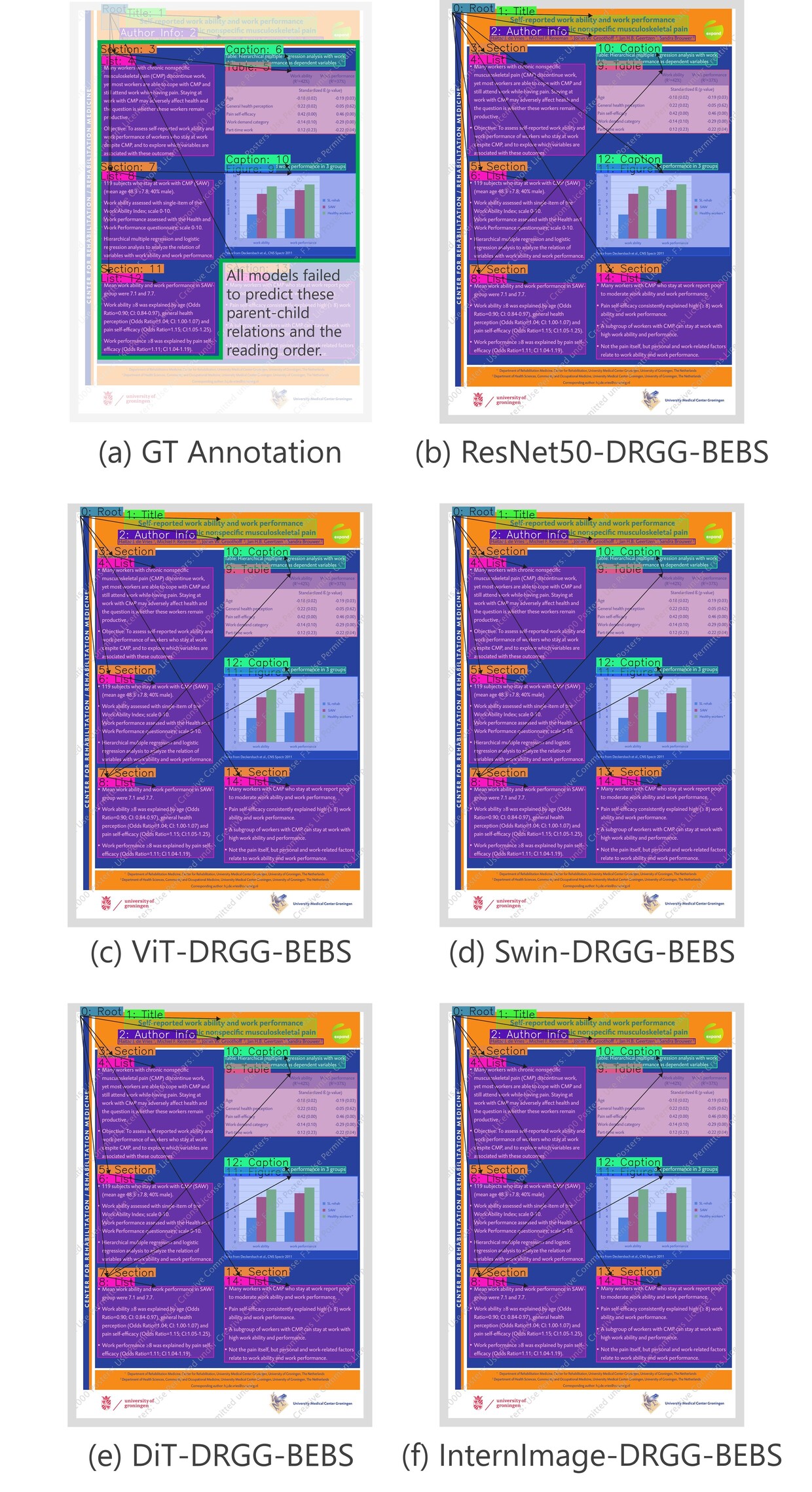}
\caption{
Example of a poster with the GT annotation and the predicted trees.
The predicted trees received low STEDS (46.67).
}
\label{fig:error_116793_all}
\end{figure*}

\begin{figure*}[t!]
\centering
\includegraphics[width=\linewidth,height=\textheight,keepaspectratio]{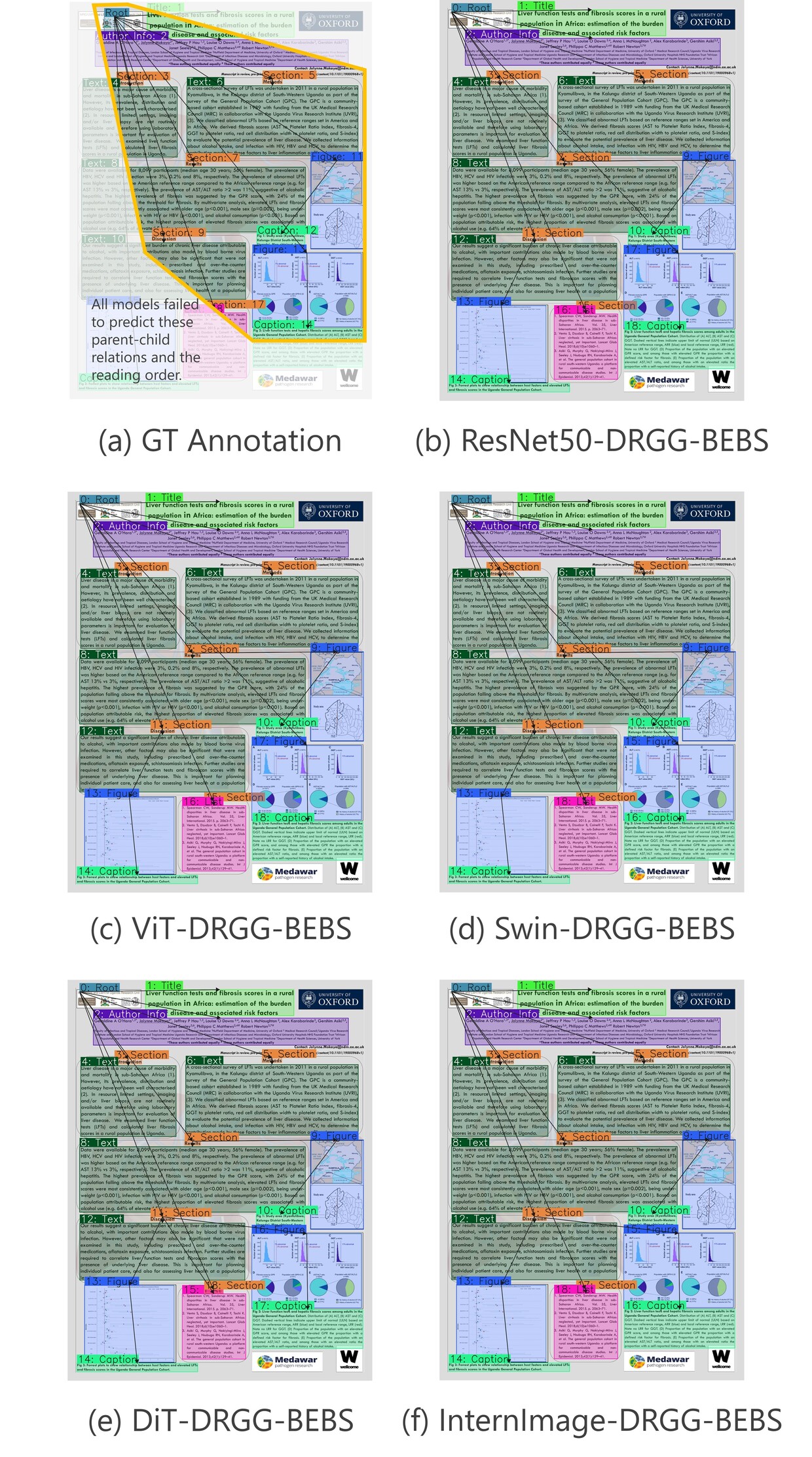}
\caption{
Example of a poster with the GT annotation and the predicted trees.
The predicted trees received low STEDS (53.68).
}
\label{fig:error_15758_all}
\end{figure*}

Figures~\ref{fig:error_121663_all}--\ref{fig:error_15758_all} present additional challenging examples as a supplement to Fig.~\ref{fig:error_example}.
In all cases, the model fails to predict inter-element relations due to irregular spatial arrangements.
Accurate prediction of such relations requires capturing semantic grouping and structural plausibility over longer sequences.

\clearpage
\clearpage

\subsection{Supplement to Analysis of Reading Order Prediction Improvements by Beam Search}

Tables~\ref{tab:ro_direction_accuracy_all}--\ref{tab:ro_distance_accuracy_all} extend Tables~\ref{tab:ro_direction_accuracy}--\ref{tab:ro_distance_accuracy} by including the results on all backbones and decoders.
These results are consistent with the conclusion drawn from Tables~\ref{tab:ro_direction_accuracy}--\ref{tab:ro_distance_accuracy}.

\begin{table*}[t!]
\centering
\small
\caption{
Per-direction accuracy of the reading order prediction with DRGG-BS or DRGG-BEBS.
Values in parentheses indicate the accuracy improvement over DRGG or DRGG-BE.
${\star}$ and ${\star\star}$ indicate significant improvements over DRGG or DRGG-BE at $p < 0.005$ and $p < 0.05$, respectively, according to McNemar's test.
}
\resizebox{\linewidth}{!}{
\begin{tabular}{ll|rrrrrrrr}
\toprule
\multicolumn{1}{l}{Backbone} & \multicolumn{1}{l}{Decoder} & \multicolumn{1}{|l}{Right} & \multicolumn{1}{l}{Bottom-Right} & \multicolumn{1}{l}{Bottom} & \multicolumn{1}{l}{Bottom-Left} & \multicolumn{1}{l}{Left} & \multicolumn{1}{l}{Top-Left} & \multicolumn{1}{l}{Top} & \multicolumn{1}{l}{Top-Right} \\
\midrule
\multirow{2}{*}{ResNet-50} & DRGG-BS & 79.4 (+5.0)$^{\star}$ & 91.5 (+1.7)$^{\star\star}$ & 93.9 (+4.0)$^{\star}$ & 87.7 (+6.1)$^{\star}$ & 77.8 (+6.5)$^{\star}$ & 58.3 (+0.0) & 73.6 (+5.8)$^{\star}$ & 67.4 (+8.2)$^{\star}$ \\
    & DRGG-BEBS & 86.0 (+4.0)$^{\star}$ & 94.5 (+0.6) & 96.7 (+2.0)$^{\star}$ & 92.7 (+2.4)$^{\star}$ & 84.2 (+3.6)$^{\star}$ & 73.3 (-1.7) & 87.5 (+2.9) & 79.8 (+5.5)$^{\star}$ \\
\midrule
\multirow{2}{*}{ViT} & DRGG-BS & 84.6 (+2.9)$^{\star}$ & 93.6 (+1.2)$^{\star\star}$ & 95.3 (+1.8)$^{\star}$ & 91.3 (+2.6)$^{\star}$ & 82.8 (+2.1)$^{\star}$ & 63.3 (+5.0) & 74.5 (+4.1)$^{\star}$ & 78.0 (+5.2)$^{\star}$ \\
    & DRGG-BEBS & 87.7 (+2.8)$^{\star}$ & 95.2 (+0.0) & 97.2 (+1.3)$^{\star}$ & 93.1 (+2.0)$^{\star}$ & 85.4 (+2.6)$^{\star}$ & 75.0 (+5.0) & 89.3 (+2.3) & 81.4 (+5.7)$^{\star}$ \\
\midrule
\multirow{2}{*}{Swin} & DRGG-BS & 85.6 (+4.0)$^{\star}$ & 93.9 (+1.3)$^{\star}$ & 95.2 (+1.4)$^{\star}$ & 91.7 (+2.6)$^{\star}$ & 83.5 (+3.2)$^{\star}$ & 60.0 (+5.0) & 77.1 (+4.1)$^{\star\star}$ & 77.2 (+4.7)$^{\star}$ \\
    & DRGG-BEBS & 87.9 (+2.4)$^{\star}$ & 95.0 (-0.5) & 97.0 (+1.1)$^{\star}$ & 92.7 (+1.0)$^{\star\star}$ & 85.5 (+2.8)$^{\star}$ & 73.3 (-5.0) & 89.0 (+2.0) & 81.1 (+3.7)$^{\star}$ \\
\midrule
\multirow{2}{*}{DiT} & DRGG-BS & 82.4 (+3.1)$^{\star}$ & 91.7 (+0.8) & 94.5 (+1.8)$^{\star}$ & 90.1 (+3.4)$^{\star}$ & 81.6 (+3.8)$^{\star}$ & 50.0 (-5.0) & 69.6 (+2.9) & 75.8 (+5.9)$^{\star}$ \\
    & DRGG-BEBS & 85.3 (+2.8)$^{\star}$ & 94.6 (+0.5) & 96.6 (+1.6)$^{\star}$ & 92.9 (+2.7)$^{\star}$ & 83.3 (+3.3)$^{\star}$ & 71.7 (-3.3) & 88.1 (+2.3) & 79.3 (+5.5)$^{\star}$ \\
\midrule
\multirow{2}{*}{InternImage} & DRGG-BS & 85.1 (+2.6)$^{\star}$ & 93.4 (+0.2) & 95.8 (+1.5)$^{\star}$ & 92.1 (+2.4)$^{\star}$ & 83.4 (+2.1)$^{\star}$ & 63.3 (+3.3) & 80.9 (+5.2)$^{\star}$ & 77.8 (+4.7)$^{\star}$ \\
    & DRGG-BEBS & 87.5 (+2.1)$^{\star}$ & 94.8 (+0.0) & 96.4 (+0.7)$^{\star}$ & 92.9 (+1.3)$^{\star}$ & 85.9 (+2.9)$^{\star}$ & 73.3 (+3.3) & 88.1 (+2.3)$^{\star\star}$ & 80.1 (+3.0)$^{\star}$ \\
\bottomrule
\end{tabular}
}
\label{tab:ro_direction_accuracy_all}
\end{table*}

\begin{table*}[t!]
\centering
\small
\caption{
Per-distance accuracy of the reading order prediction.
The format and experimental settings follow Table~\ref{tab:ro_direction_accuracy_all}.
}
\begin{tabular}{ll|rrrrrr}
\toprule
Backbone & Decoder & (0, 1] & (1, 2] & (2, 4] & (4, 8] & (8, 16] & (16, $\infty$) \\
\midrule
\multirow{2}{*}{ResNet-50} & DRGG-BS & 91.3\ (+4.1)$^{\star}$ & 76.5\ (+5.9)$^{\star}$ & 62.1\ (+6.7)$^{\star}$ & 71.7\ (+13.1)$^{\star}$ & 71.0\ (+9.9)$^{\star}$ & 78.9\ (+8.1)$^{\star\star}$ \\
    & DRGG-BEBS & 94.7\ (+1.9)$^{\star}$ & 84.0\ (+4.1)$^{\star}$ & 77.4\ (+6.4)$^{\star}$ & 85.4\ (+5.6)$^{\star\star}$ & 86.3\ (+5.3)$^{\star\star}$ & 89.4\ (+3.3) \\
\midrule
\multirow{2}{*}{ViT} & DRGG-BS & 93.2\ (+2.0)$^{\star}$ & 82.6\ (+2.7)$^{\star}$ & 72.1\ (+4.4)$^{\star}$ & 83.8\ (+8.6)$^{\star}$ & 77.1\ (+3.1) & 85.4\ (+3.3) \\
    & DRGG-BEBS & 95.3\ (+1.3)$^{\star}$ & 85.5\ (+3.1)$^{\star}$ & 80.5\ (+4.4)$^{\star}$ & 86.4\ (+8.1)$^{\star}$ & 87.0\ (+2.3) & 89.4\ (+4.9)$^{\star\star}$ \\
\midrule
\multirow{2}{*}{Swin} & DRGG-BS & 93.3\ (+2.0)$^{\star}$ & 83.2\ (+3.1)$^{\star}$ & 74.1\ (+4.7)$^{\star}$ & 81.3\ (+6.1)$^{\star\star}$ & 76.3\ (+3.8) & 87.8\ (+4.9) \\
    & DRGG-BEBS & 95.2\ (+1.0)$^{\star}$ & 85.5\ (+2.6)$^{\star}$ & 78.7\ (+2.2) & 84.3\ (+3.0) & 87.0\ (+1.5) & 90.2\ (+4.9)$^{\star\star}$ \\
\midrule
\multirow{2}{*}{DiT} & DRGG-BS & 91.8\ (+2.0)$^{\star}$ & 80.6\ (+4.1)$^{\star}$ & 72.5\ (+6.0)$^{\star}$ & 77.3\ (+4.0) & 74.8\ (+3.1) & 86.2\ (+2.4) \\
    & DRGG-BEBS & 94.4\ (+1.6)$^{\star}$ & 84.1\ (+3.8)$^{\star}$ & 77.2\ (+5.1)$^{\star}$ & 83.3\ (+3.5) & 87.8\ (+3.8) & 91.1\ (+0.8) \\
\midrule
\multirow{2}{*}{InternImage} & DRGG-BS & 93.7\ (+1.5)$^{\star}$ & 83.3\ (+3.6)$^{\star}$ & 74.9\ (+3.1) & 84.8\ (+6.1)$^{\star}$ & 76.3\ (+1.5) & 87.0\ (+5.7)$^{\star\star}$ \\
    & DRGG-BEBS & 94.8\ (+0.9)$^{\star}$ & 85.1\ (+2.1)$^{\star}$ & 78.0\ (+4.0)$^{\star}$ & 88.9\ (+2.5) & 86.3\ (+2.3) & 87.8\ (+3.3) \\
\bottomrule
\end{tabular}

\label{tab:ro_distance_accuracy_all}
\end{table*}
\subsection{Supplement to Analysis of Parent-Child Prediction Improvements by BBox Embedding}

Tables~\ref{tab:pc_direction_accuracy_all}--\ref{tab:pc_distance_accuracy_all} extend Tables~\ref{tab:pc_direction_accuracy}--\ref{tab:pc_distance_accuracy} by including the results on all backbones and decoders.
These results are consistent with the conclusion drawn from Tables~\ref{tab:pc_direction_accuracy}--\ref{tab:pc_distance_accuracy}.


\begin{table*}[t!]
\centering
\small
\caption{
Per-direction accuracy of the parent-child prediction with DRGG-BE or DRGG-BEBS.
Values in parentheses indicate the accuracy improvement over DRGG or DRGG-BS.
${\star}$ and ${\star\star}$ indicate significant improvements over DRGG or DRGG-BS at $p < 0.005$ and $p < 0.05$, respectively, according to McNemar's test.
}
\resizebox{\linewidth}{!}{
\begin{tabular}{ll|rrrrrrrr}
\toprule
\multicolumn{1}{l}{Backbone} & \multicolumn{1}{l}{Decoder} & \multicolumn{1}{|l}{Right} & \multicolumn{1}{l}{Bottom-Right} & \multicolumn{1}{l}{Bottom} & \multicolumn{1}{l}{Bottom-Left} & \multicolumn{1}{l}{Left} & \multicolumn{1}{l}{Top-Left} & \multicolumn{1}{l}{Top} & \multicolumn{1}{l}{Top-Right} \\
\midrule
\multirow{2}{*}{ResNet-50} & DRGG-BE & 89.2\ (+18.2)$^{\star}$ & 91.4\ (+12.1)$^{\star}$ & 95.7\ (+11.5)$^{\star}$ & 90.2\ (+19.5)$^{\star}$ & 69.5\ (+16.3)$^{\star}$ & 88.4 (+18.6)$^{\star\star}$ & 86.5\ (+18.0)$^{\star}$ & 82.4\ (+22.8)$^{\star}$ \\
    & DRGG-BEBS & 93.3\ (+13.3)$^{\star}$ & 95.3\ (+8.4)$^{\star}$ & 98.0\ (+9.2)$^{\star}$ & 92.6\ (+13.6)$^{\star}$ & 74.5\ (+19.1)$^{\star}$ & 90.7 (+18.6)$^{\star\star}$ & 88.8\ (+14.0)$^{\star}$ & 83.4 (+10.4)$^{\star\star}$ \\
\midrule
\multirow{2}{*}{ViT} & DRGG-BE & 91.3\ (+5.2)$^{\star}$ & 94.2\ (+4.8)$^{\star}$ & 96.6\ (+5.9)$^{\star}$ & 91.7\ (+11.0)$^{\star}$ & 75.2\ (+12.1)$^{\star}$ & 83.7 (+9.3) & 89.9\ (+26.4)$^{\star}$ & 83.4\ (+15.0)$^{\star}$ \\
    & DRGG-BEBS & 94.3\ (+5.3)$^{\star}$ & 97.1\ (+3.7)$^{\star}$ & 98.4\ (+5.5)$^{\star}$ & 94.7\ (+9.3)$^{\star}$ & 77.3 (+7.1) & 86.0 (+7.0) & 89.9\ (+20.8)$^{\star}$ & 87.6\ (+12.4)$^{\star}$ \\
\midrule
\multirow{2}{*}{Swin} & DRGG-BE & 92.0\ (+9.3)$^{\star}$ & 94.6\ (+6.1)$^{\star}$ & 96.9\ (+7.2)$^{\star}$ & 89.4\ (+9.1)$^{\star}$ & 70.2 (+6.4)$^{\star\star}$ & 88.4\ (+20.9)$^{\star}$ & 91.6\ (+24.7)$^{\star}$ & 90.7\ (+20.7)$^{\star}$ \\
    & DRGG-BEBS & 94.8\ (+7.9)$^{\star}$ & 97.1\ (+6.1)$^{\star}$ & 98.5\ (+6.6)$^{\star}$ & 92.1\ (+8.5)$^{\star}$ & 73.0\ (+9.9)$^{\star}$ & 88.4 (+14.0)$^{\star\star}$ & 91.0\ (+21.3)$^{\star}$ & 90.2\ (+14.0)$^{\star}$ \\
\midrule
\multirow{2}{*}{DiT} & DRGG-BE & 89.2\ (+7.6)$^{\star}$ & 93.2\ (+7.0)$^{\star}$ & 96.0\ (+7.4)$^{\star}$ & 91.1\ (+11.3)$^{\star}$ & 73.0 (+9.9)$^{\star\star}$ & 88.4\ (+25.6)$^{\star}$ & 88.2\ (+34.8)$^{\star}$ & 78.2\ (+9.8)$^{\star}$ \\
    & DRGG-BEBS & 92.8\ (+7.7)$^{\star}$ & 96.3\ (+6.5)$^{\star}$ & 97.8\ (+7.2)$^{\star}$ & 94.7\ (+12.1)$^{\star}$ & 80.1\ (+14.9)$^{\star}$ & 83.7 (+23.3)$^{\star\star}$ & 90.4\ (+32.6)$^{\star}$ & 84.5\ (+9.3)$^{\star}$ \\
\midrule
\multirow{2}{*}{InternImage} & DRGG-BE & 89.9\ (+5.2)$^{\star}$ & 94.2\ (+5.5)$^{\star}$ & 97.2\ (+6.7)$^{\star}$ & 91.1\ (+10.2)$^{\star}$ & 70.9 (+4.3) & 83.7 (+11.6) & 87.6\ (+16.3)$^{\star}$ & 89.1\ (+19.2)$^{\star}$ \\
    & DRGG-BEBS & 93.8\ (+6.1)$^{\star}$ & 96.9\ (+5.5)$^{\star}$ & 98.5\ (+6.1)$^{\star}$ & 94.3\ (+10.8)$^{\star}$ & 75.9 (+7.1)$^{\star\star}$ & 88.4 (+11.6) & 90.4\ (+13.5)$^{\star}$ & 92.2\ (+16.6)$^{\star}$ \\
\bottomrule
\end{tabular}
}
\label{tab:pc_direction_accuracy_all}
\end{table*}

\begin{table*}[t!]
\centering
\small
\caption{
Per-distance accuracy of the parent-child prediction.
The format and experimental settings follow Table~\ref{tab:pc_direction_accuracy_all}.
}
\begin{tabular}{ll|rrrrrr}
\toprule
Backbone & Decoder & (0, 1] & (1, 2] & (2, 4] & (4, 8] & (8, 16] & (16, $\infty$) \\
\midrule
\multirow{2}{*}{ResNet-50} & DRGG-BE & 96.3\ (+7.7)$^{\star}$ & 88.0\ (+17.4)$^{\star}$ & 86.2\ (+22.1)$^{\star}$ & 88.7\ (+21.4)$^{\star}$ & 86.4\ (+22.5)$^{\star}$ & 88.3\ (+32.2)$^{\star}$ \\
    & DRGG-BEBS & 97.8\ (+6.4)$^{\star}$ & 92.2\ (+13.1)$^{\star}$ & 92.0\ (+14.7)$^{\star}$ & 92.3\ (+13.3)$^{\star}$ & 91.1\ (+16.5)$^{\star}$ & 93.7\ (+28.3)$^{\star}$ \\
\midrule
\multirow{2}{*}{ViT} & DRGG-BE & 97.2\ (+5.7)$^{\star}$ & 89.8\ (+4.7)$^{\star}$ & 90.7\ (+8.7)$^{\star}$ & 91.8\ (+8.5)$^{\star}$ & 87.0\ (+8.2)$^{\star}$ & 91.2\ (+23.9)$^{\star}$ \\
    & DRGG-BEBS & 98.3\ (+5.1)$^{\star}$ & 92.8\ (+4.2)$^{\star}$ & 94.8\ (+6.6)$^{\star}$ & 94.9\ (+5.7)$^{\star}$ & 94.0\ (+9.5)$^{\star}$ & 95.6\ (+23.4)$^{\star}$ \\
\midrule
\multirow{2}{*}{Swin} & DRGG-BE & 97.2\ (+5.6)$^{\star}$ & 90.5\ (+8.5)$^{\star}$ & 91.4\ (+10.7)$^{\star}$ & 90.7\ (+9.7)$^{\star}$ & 90.8\ (+15.8)$^{\star}$ & 91.7\ (+26.8)$^{\star}$ \\
    & DRGG-BEBS & 98.5\ (+5.8)$^{\star}$ & 93.4\ (+7.6)$^{\star}$ & 94.2\ (+9.0)$^{\star}$ & 93.5\ (+7.9)$^{\star}$ & 93.4\ (+12.3)$^{\star}$ & 94.6\ (+20.5)$^{\star}$ \\
\midrule
\multirow{2}{*}{DiT} & DRGG-BE & 96.5\ (+7.5)$^{\star}$ & 88.1\ (+6.6)$^{\star}$ & 88.7\ (+10.8)$^{\star}$ & 90.5\ (+8.8)$^{\star}$ & 88.6\ (+14.9)$^{\star}$ & 87.8\ (+15.6)$^{\star}$ \\
    & DRGG-BEBS & 97.9\ (+7.2)$^{\star}$ & 92.9\ (+7.6)$^{\star}$ & 92.4\ (+9.5)$^{\star}$ & 93.1\ (+7.7)$^{\star}$ & 93.0\ (+15.2)$^{\star}$ & 94.6\ (+20.0)$^{\star}$ \\
\midrule
\multirow{2}{*}{InternImage} & DRGG-BE & 97.0\ (+4.8)$^{\star}$ & 90.3\ (+7.1)$^{\star}$ & 90.0\ (+8.9)$^{\star}$ & 91.5\ (+8.7)$^{\star}$ & 89.9\ (+12.7)$^{\star}$ & 91.7\ (+23.9)$^{\star}$ \\
    & DRGG-BEBS & 98.4\ (+5.2)$^{\star}$ & 93.1\ (+7.1)$^{\star}$ & 93.5\ (+8.0)$^{\star}$ & 94.6\ (+7.5)$^{\star}$ & 95.6\ (+13.3)$^{\star}$ & 95.6\ (+17.6)$^{\star}$ \\
\bottomrule
\end{tabular}
\label{tab:pc_distance_accuracy_all}
\end{table*}



\subsection{Effect of Beam Width}

\begin{figure}[t!]
\centering
\includegraphics[width=\columnwidth]{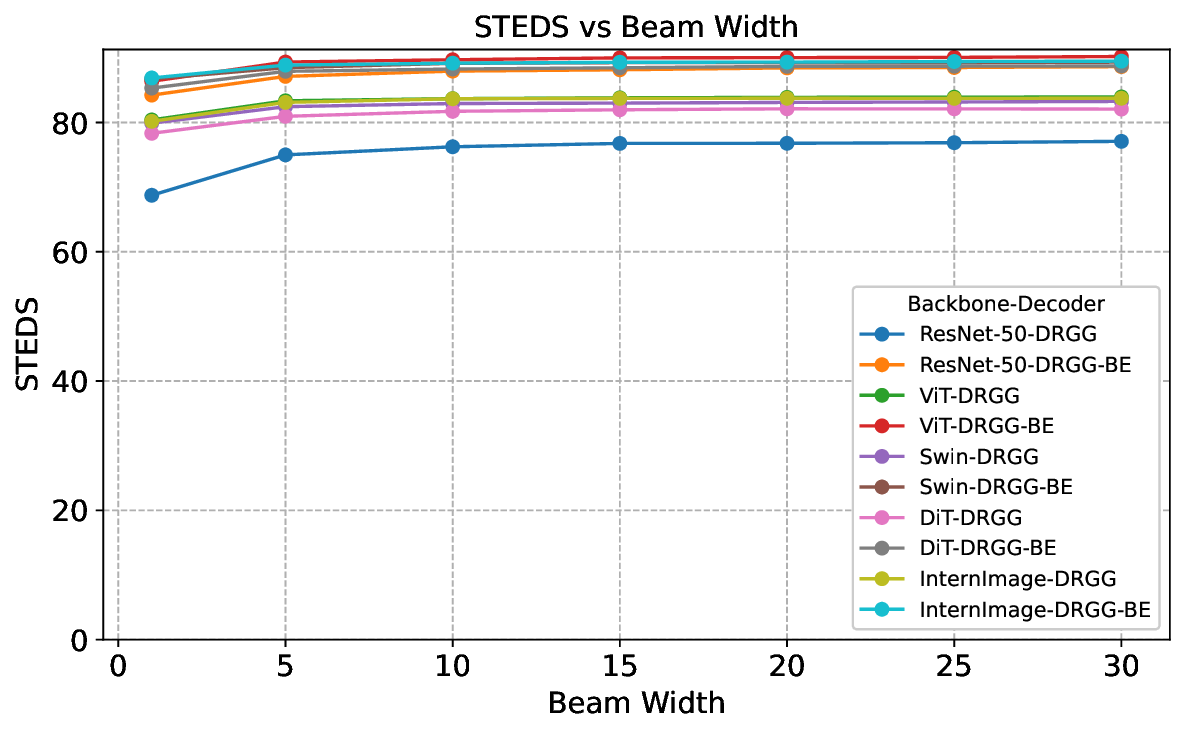}
\caption{Effect of beam width on STEDS performance}
\label{fig:steds_beamwidth}
\end{figure}

\begin{figure}[t!]
\centering
\includegraphics[width=\columnwidth]{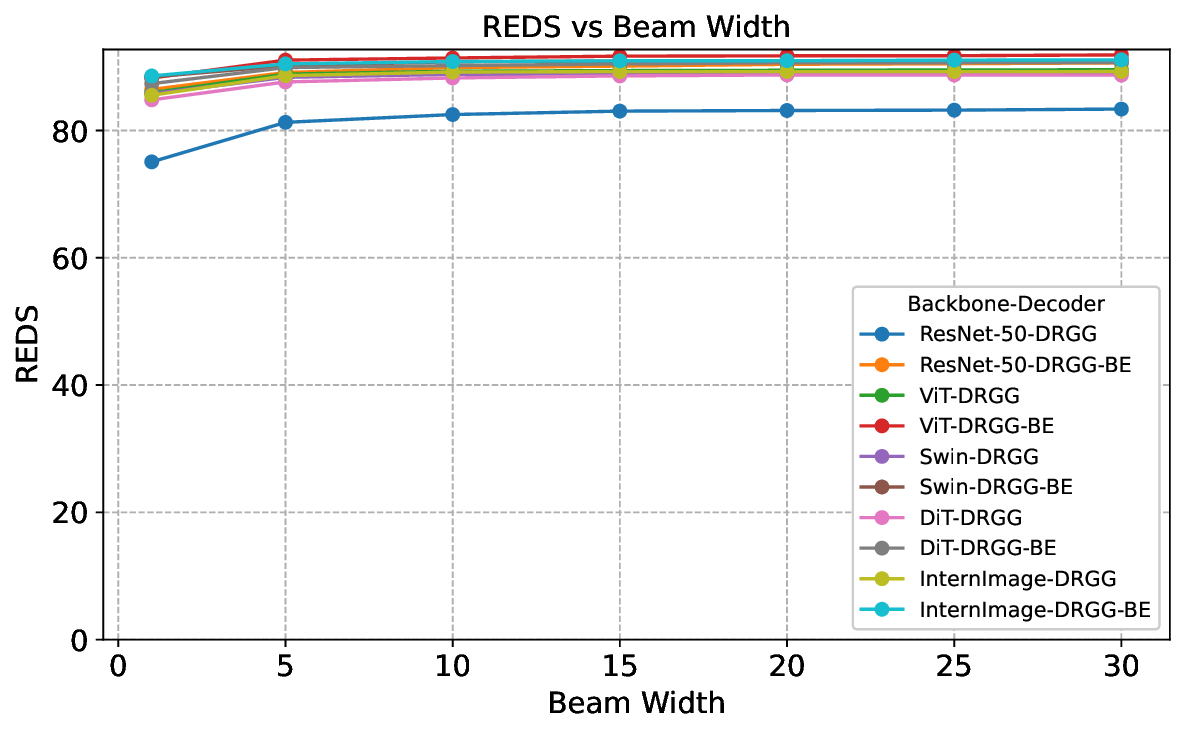}
\caption{Effect of beam width on REDS performance}
\label{fig:reds_beamwidth}
\end{figure}

\begin{figure}[t!]
\centering
\includegraphics[width=\columnwidth]{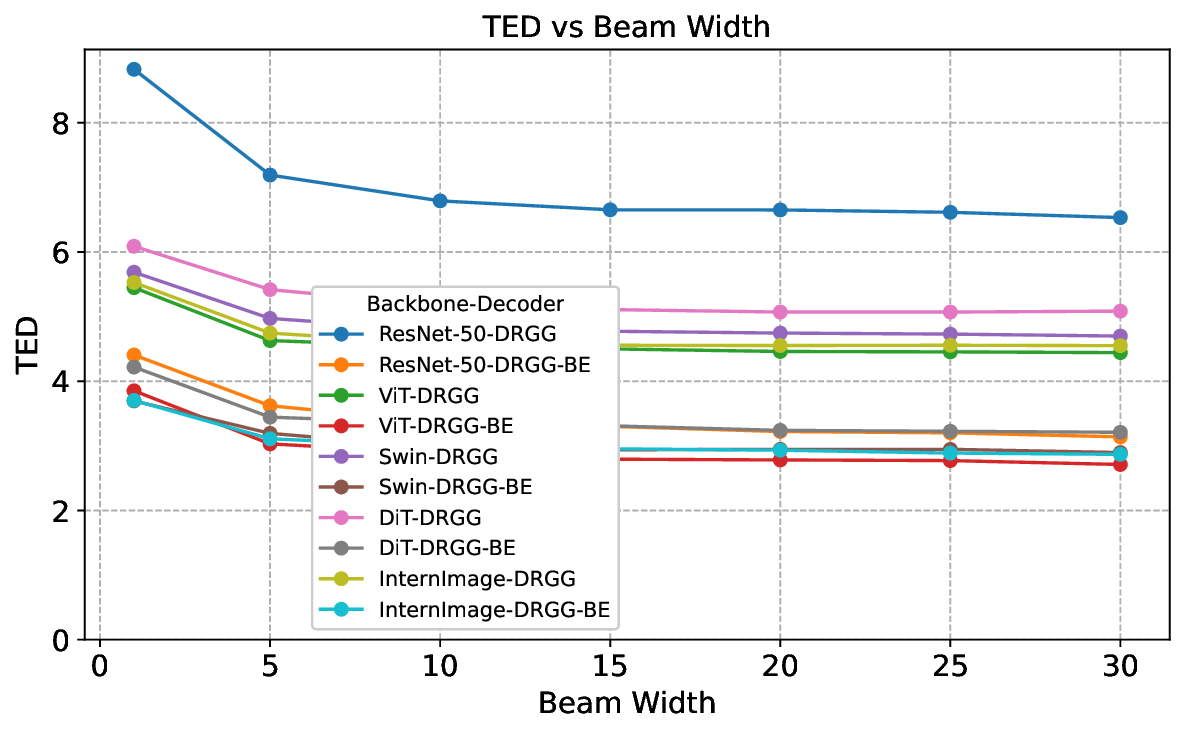}
\caption{Effect of beam width on TED performance}
\label{fig:ted_beamwidth}
\end{figure}

Figures~\ref{fig:steds_beamwidth}--\ref{fig:ted_beamwidth} show the performance variations of each evaluation metric when the beam width is varied among \{1, 5, 10, 15, 20, 25, 30\}.  
Across all metrics, the performance saturates around a beam width of 15 to 20.  
This behavior can be attributed to the fact that the average number of BBoxes in SciPostLayoutTree is approximately 25.
Increasing the beam width beyond this point results in only marginal performance improvements.

\subsection{Histogram of Scores}

\begin{figure*}[t!]
\centering
\includegraphics[width=\linewidth]{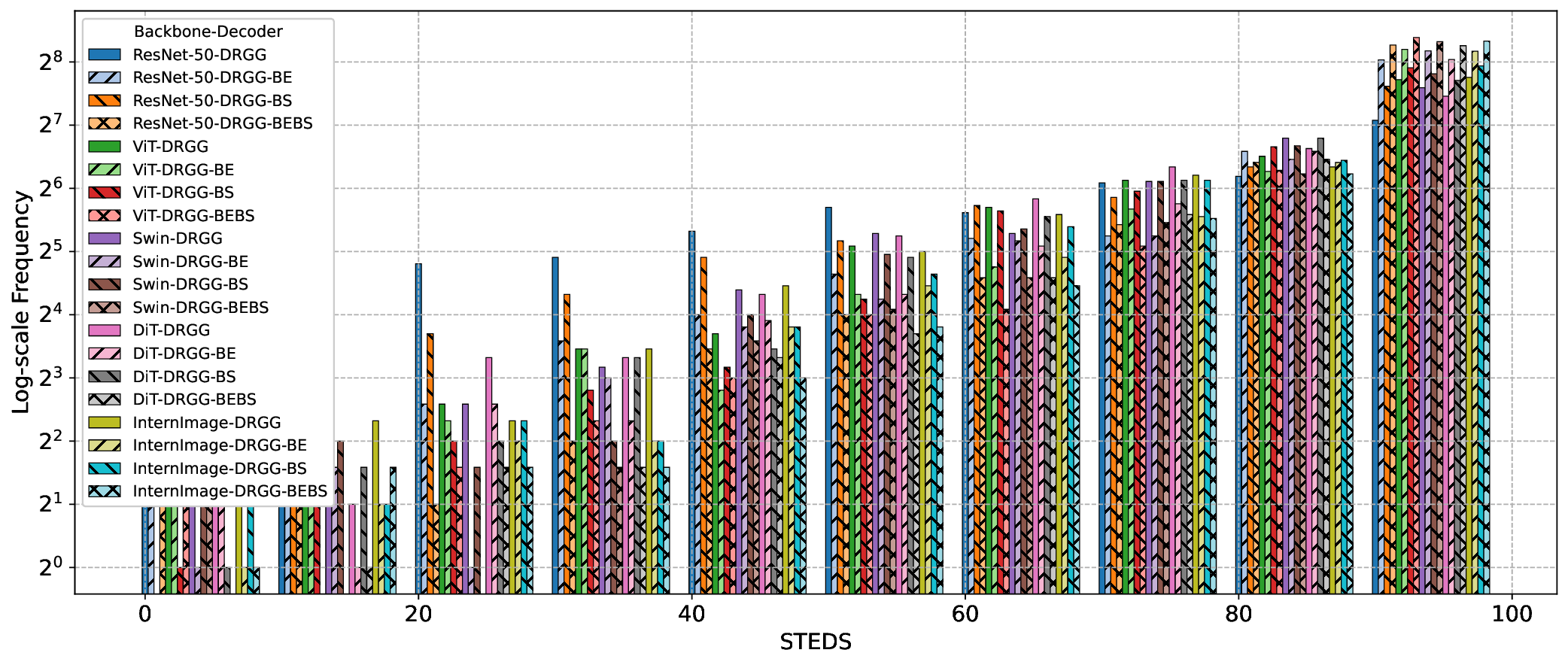}
\caption{Histogram of STEDS with a bin width of 10}
\label{fig:steds_histogram}
\end{figure*}

\begin{figure*}[t!]
\centering
\includegraphics[width=\linewidth]{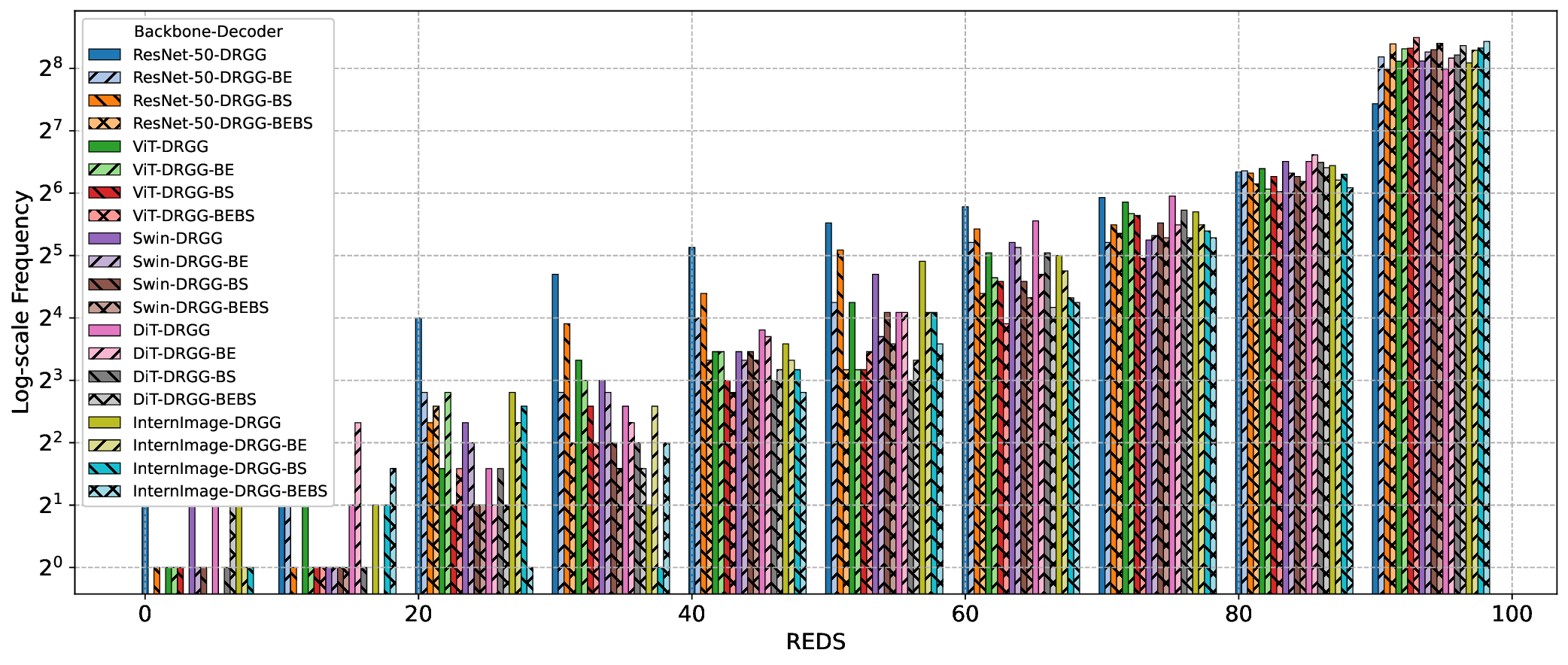}
\caption{Histogram of REDS with a bin width of 10}
\label{fig:reds_histogram}
\end{figure*}

\begin{figure*}[t!]
\centering
\includegraphics[width=\linewidth]{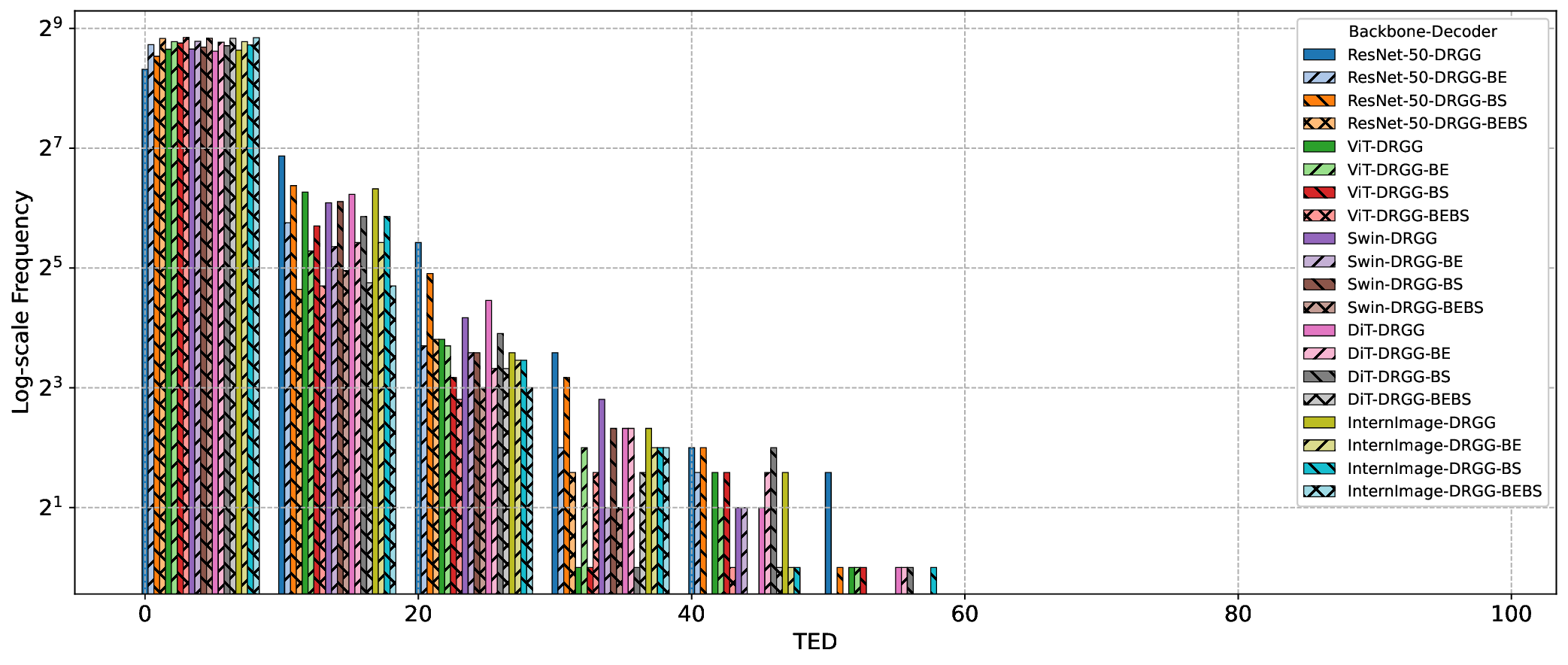}
\caption{Histogram of TED with a bin width of 10}
\label{fig:ted_histogram}
\end{figure*}

Figures~\ref{fig:steds_histogram}--\ref{fig:ted_histogram} show histograms of each evaluation metric with a bin width of 10.  
Although the models often decode trees similar to the GT trees, low scores are still observed in a non-negligible number of cases.
These results indicate that the models remain suboptimal.

\end{document}